\documentclass{article}

\usepackage{microtype}
\usepackage{graphicx}
\usepackage{subcaption}
\usepackage{booktabs} 

\usepackage{wrapfig}
\usepackage{amsmath}
\usepackage{amsfonts}       
\newcommand{\bx}{\mathbf{x}}

\newcommand{\bz}{\mathbf{z}}

\newcommand{\RE}{\ensuremath{\mathbb{R}}}
\usepackage{enumitem}
\usepackage{textcomp}
\usepackage{multirow}

\DeclareGraphicsExtensions{.pdf,.png}

\usepackage{hyperref}

\usepackage[accepted]{icml2021}

\icmltitlerunning{Autoencoding Under Normalization Constraints}

\begin{document}

\twocolumn[
\icmltitle{Autoencoding Under Normalization Constraints}



\icmlsetsymbol{equal}{*}

\begin{icmlauthorlist}
\icmlauthor{Sangwoong Yoon}{snu}
\icmlauthor{Yung-Kyun Noh}{hanyang,kias}
\icmlauthor{Frank C. Park}{snu,saige}
\end{icmlauthorlist}

\icmlaffiliation{snu}{Department of Mechanical Engineering, Seoul National University, Seoul, Republic of Korea}
\icmlaffiliation{hanyang}{Department of Computer Science, Hanyang University, Seoul, Republic of Korea}
\icmlaffiliation{kias}{Korea Institute of Advanced Studies, Seoul, Republic of Korea}
\icmlaffiliation{saige}{Saige Research, Seoul, Republic of Korea}

\icmlcorrespondingauthor{Yung-Kyun Noh}{nohyung@hanyang.ac.kr}
\icmlcorrespondingauthor{Frank C. Park}{fcp@snu.ac.kr}

\icmlkeywords{Autoencoders, Out-of-distribution detection, Energy-based models}

\vskip 0.3in
]



\printAffiliationsAndNotice{}  

\begin{abstract}
Likelihood is a standard estimate for outlier detection. The specific role of the normalization constraint is to ensure that the out-of-distribution (OOD) regime has a small likelihood when samples are learned using maximum likelihood. Because autoencoders do not possess such a process of normalization, they often fail to recognize outliers even when they are obviously OOD. We propose the Normalized Autoencoder (NAE), a normalized probabilistic model constructed from an autoencoder. The probability density of NAE is defined using the reconstruction error of an autoencoder, which is differently defined in the conventional energy-based model. In our model, normalization is enforced by suppressing the reconstruction of negative samples, significantly improving the outlier detection performance. Our experimental results confirm the efficacy of NAE, both in detecting outliers and in generating in-distribution samples.
\end{abstract}

\section{Introduction}

An autoencoder \citep{rumelhart1986} is a neural network trained to 
reconstruct samples from a training data distribution. Since in principle
the quality of reconstruction is expected to be poor for inputs that
deviate significantly from the training data, autoencoders are widely
used in outlier detection \citep{japkowicz1995novelty}, in which an input
with a large reconstruction error is classified as out-of-distribution
(OOD).  Autoencoders for outlier detection have been applied
in domains ranging from video surveillance \citep{zhao2017spatio} to medical diagnosis \citep{lu2018anomaly}.

\begin{figure}
    \centering
    \includegraphics[width=0.48\textwidth]{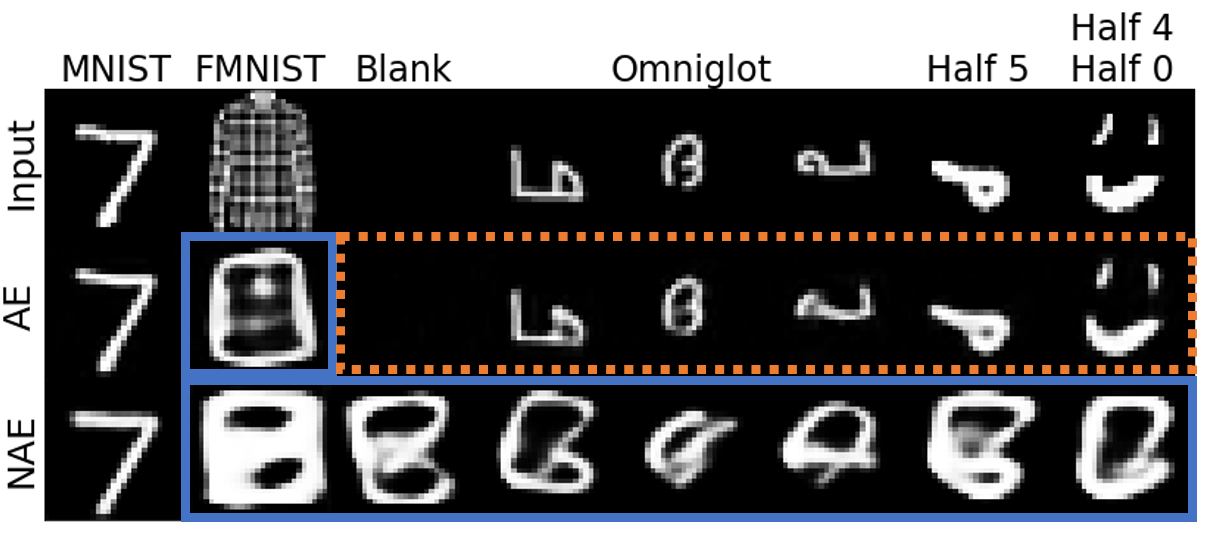}
    \vskip -0.1in
    \caption{Examples of reconstructed outliers. The last two rows show the reconstructions from a conventional autoencoder (AE) and NAE. Both autoencoders are trained on MNIST, and other inputs are outliers. The architecture of the two autoencoders is identical.
    Successful detection of an outlier is highlighted with blue solid rectangles, while detection failures due to the reconstruction of outliers are denoted with an orange dotted rectangle. Note that AE is not the identity mapping, as it fails to reconstruct the shirt. }
    \label{fig:mnist_recon}
    \vskip -0.2in
\end{figure}

However, autoencoders have been known to reconstruct outliers consistently across a wide range of experimental settings \cite{lyudchik2016outlier, tong2019fixing, zong2018deep, gong2019memorizing}. 
We name this phenomenon \emph{outlier reconstruction}.
Figure \ref{fig:mnist_recon} shows examples of some outliers reconstructed by an autoencoder trained
with MNIST data; the autoencoder is able to reconstruct a wide range
of OOD inputs, including constant black pixels, Omniglot characters,
and fragments of MNIST digits.
The early works on regularized autoencoders \cite{vincent2008extracting,rifai2011contractive,ng2011sparse} focus for the most part on preventing the autoencoder from turning into the identity mapping that reconstructs every input.
Nonetheless, outlier reconstruction can still occur even when the autoencoder is not the identity as shown by the non-identity autoencoder in Figure \ref{fig:mnist_recon}.
Not surprisingly, outlier reconstruction is a leading cause of autoencoder's detection failure.

On the other hand, in a normalized probabilistic model, it is known that maximum likelihood
learning suppresses the assignment of probability mass in OOD regions in
order to keep the model normalized. 
Thus, the likelihood is widely used as a predictor for outlier detection \cite{bishop1994novelty}.
Meanwhile, an autoencoder is not a probabilistic model of the data and does not have a suppression mechanism corresponding to the normalization in other probabilistic models.
As a result, the reconstruction of outliers are not inhibited during training of an autoencoder.

This paper formulates an autoencoder as a normalized probabilistic model to introduce a mechanism for preventing outlier reconstruction.
In our formulation, which we call the \textbf{Normalized Autoencoder (NAE)}, the reconstruction error is re-interpreted as an energy function, i.e., the unnormalized negative log-density, and defines a probabilistic model from an autoencoder.
During maximum likelihood learning of NAE, outlier reconstruction is naturally suppressed by enforcing the normalization constraint, and the resulting autoencoder is significantly less prone to reconstruct outliers, as shown in Figure \ref{fig:mnist_recon}.

In each training iteration of NAE, samples generated from the model is used to update the normalization constraint which is implicitly computed as in other energy-based models.
Since running a Markov Chain Monte Carlo (MCMC) sampler until convergence every iteration is computationally infeasible, an approximate sampling strategy has to be employed.
We observe that training with popular sampling strategies such as Contrastive Divergence
(CD; \citet{hinton2002training}) and Persistent CD (PCD; \citet{tieleman2008training}) may often produce poor density estimates.
Instead, we propose \textbf{on-manifold initialization (OMI)}, a method of initializing an MCMC chain on manifold defined by the decoder of an autoencoder.
OMI selects high-model-density initial states by leveraging the assumption that points on the decoder manifold typically have small reconstruction error, i.e. high model density.
With OMI, NAE can accurately recover the data density and thus become an effective outlier detector.

Intriguingly, although technically a normalized probabilistic model,
the variational autoencoder (VAE; \citet{kingma2013auto}) also 
reconstructs outliers and assigns a spuriously high likelihood on
OOD data \cite{nalisnick2018do,xiao2020likelihood} for reasons that are as-yet unclear.

Our main contributions can be summarized as follows:
\begin{itemize}[leftmargin=1em,topsep=0pt,noitemsep]
\item We propose NAE, a novel generative model constructed from an autoencoder;
\item We propose OMI, a sampling strategy tailored for NAE;
\item We empirically show that NAE is highly effective for outlier detection and can perform other generative tasks.
\end{itemize}

Section 2 provides brief background on autoencoders and energy-based models. NAE is described in Section 3, and OMI is described in Section 4.
Related works are reviewed in Section 5. Section 6 presents experimental results. Section 7 provide discussions and conclude the paper.
Our source code and pre-trained models are publicly available online \footnote{\scriptsize{\url{https://github.com/swyoon/normalized-autoencoders}}}.
We also provide an interactive web demo on outlier reconstruction phenomenon of autoencoders \footnote{\scriptsize\url{https://swyoon.github.io/outlier-reconstruction/}}. In this web demo, NAE is shown to reconstruct only in-distribution images, while a vanilla autoencoder inadvertently reconstructs various outliers.

\section{Background}

\subsection{Autoencoders} \label{sec:ae}

Autoencoders are neural networks trained to reconstruct an input datum  $\bx \in \mathcal{X} \subset \RE^{D_\bx}$. For an input $\bx$, the quality of its reconstruction is measured in reconstruction error $l_\theta(\bx)$, where $\theta$ denotes parameters in an autoencoder. The loss function of an autoencoder $L_\text{AE}$ for training is the expected reconstruction error of training data. 
Gradient descent training is performed via computing the gradient of $L$ with respect to model parameters $\theta$:
\begin{align}
    L_{\text{AE}} &=\mathbb{E}_{\bx \sim p(\bx)}[l_\theta(\bx)] \label{eq:ae_objective}, \\
    \nabla_\theta L_{\text{AE}} &= \mathbb{E}_{\bx \sim p(\bx)}[\nabla_\theta l_\theta(\bx)], \label{eq:ae_gradient}
\end{align}
where $\nabla_\theta$ is the gradient operator with respect to $\theta$ and $p(\bx)$ denotes the data density.

\textbf{Architecture} An autoencoder consists of two submodules, an encoder and a decoder.
An encoder $f_e(\bx): \RE^{D_\bx} \to \RE^{D_\bz}$ maps an input $\bx$ to a corresponding latent representation vector $\bz \in \mathcal{Z} \subset  \RE^{D_\bz}$, and a decoder $f_d(\bz): \RE^{D_\bz} \to \RE^{D_\bx}$ maps a latent vector $\bz$ back to the input space.
Then, the reconstruction error $l_\theta(\bx)$ is given as:
\begin{align}
    l_\theta(\bx)=\mathrm{dist}(\bx, f_d(f_e(\bx))), \label{eq:recon_error}
\end{align}
where $\mathrm{dist}(\cdot, \cdot)$ is a distance-like function measuring the deviation between an input $\bx$ and a reconstruction $f_d(f_e(\bx))$.
A typical choice is the squared $L^2$ distance, i.e., $\mathrm{dist}(\bx_1, \bx_2)=|| \bx_1 - \bx_2 ||_2^2$.
Other possible choices include $L^1$ distance, $\mathrm{dist}(\bx_1, \bx_2)=|\bx_1 - \bx_2|$, and the structural similarity (SSIM; \citet{wang2004image,bergmann2018improving}).

Note that the reconstruction error (Eq.~(\ref{eq:recon_error})) is \emph{not} a likelihood of a datum, and therefore the minimization of the reconstruction error does not correspond to the maximization of the likelihood.
Without modification, an autoencoder per se is not a probabilistic model.

\textbf{Outlier Detection} A datum is an outlier or called OOD if it lies in the $\rho$-sublevel set of a data density $\{\bx | p(\bx)\leq \rho\}$ \cite{steinwart2005classification}. 
We particularly focus on $\rho=0$, where an outlier is defined as an input from the outside of the data distribution's support. 
Most of the OOD examples which attract the attention of the research community are in fact out-of-support samples.
For example, SVHN and CIFAR-10 are out-of-support to each other, as confirmed by a supervised classifier perfectly discriminating the two datasets.
Note that the support-based definition provides invariant characterization of outliers, as no invertible transform defined on the data space alters whether a sample is in- or out-of-support.
Meanwhile, for $\rho\neq0$, the characterization of outliers are not invariant to the choice of coordinates \citet{lan2020perfect}.

In the autoencoder-based outlier detection \citep{japkowicz1995novelty}, an input is classified as OOD if its reconstruction error $l_\theta(\bx)$ is greater than a threshold $\tau$: $l_\theta(\bx)>\tau$.
The outlier reconstruction indicates that there exists an input $\bx^*$ with $p(\bx^*)\leq \rho$, but $l_\theta(\bx^*)<\tau$.
Appendix includes the detailed investigation on outlier reconstruction.

\subsection{Energy-based Models}
Unlike autoencoders, energy-based models (EBMs) are valid models for a normalized probability distribution.
The EBM represents a probability distribution through the unnormalized negative log probability, also called the energy function $E_\theta(\bx)$. Here, $\theta$ denotes the model parameters.

For a continuous input $\bx \in \mathcal{X} \subset \RE^{D_\bx}$, $E_\theta(\bx)$ defines the model density function $p_\theta(\bx)$ through Gibbs distribution:
\begin{align}
    p_\theta(\bx) = \frac{1}{\Omega_\theta} \exp(-E_\theta(\bx)/T), \label{eq:ebm}
\end{align}
where $T \in \RE^+$ is called the temperature and is often ignored by setting $T=1$.
$\Omega_\theta$ is the normalization constant and is defined as:
\begin{align}
    \Omega_\theta = \int_{\mathcal{X}} \exp(-E_\theta(\bx)/T) \mathrm{d}\bx < \infty. \label{eq:Omega}
\end{align}
The computation of $\Omega_\theta$ is usually difficult for high-dimensional $\bx$. 
However, maximum likelihood learning can still be performed without the explicit evaluation of $\Omega_\theta$.
The gradient of negative log likelihood of data is given as follows \cite{younes1999convergence}:
\begin{align}
    &\mathbb{E}_{\bx\sim p(\bx)}[-\nabla_{\theta} \log p_{\theta}(\bx)] \nonumber\\
    =& \mathbb{E}_{\bx\sim p(\bx)}[\nabla_{\theta} E_\theta(\bx)]/T + \nabla_\theta \log \Omega_\theta \label{eq:ebm1}\\
    =& \mathbb{E}_{\bx\sim p(\bx)}[\nabla_{\theta} E_\theta(\bx)]/T - \mathbb{E}_{\bx'\sim p_{\theta}(\bx)}[\nabla_{\theta} E_\theta(\bx')]/T \label{eq:ebm2}
\end{align}
$\nabla_\theta \log \Omega_\theta$ in Eq.~(\ref{eq:ebm1}) is evaluated from the energy gradients of samples $\bx'$ generated from the model in Eq.~(\ref{eq:ebm2}).
The samples from $p_\theta(\bx)$ are often called "negative" samples.
The derivation of Eq.~(\ref{eq:ebm2}) is provided in Appendix.

\begin{figure}[t]
    \centering
    \includegraphics[width=0.45\textwidth]{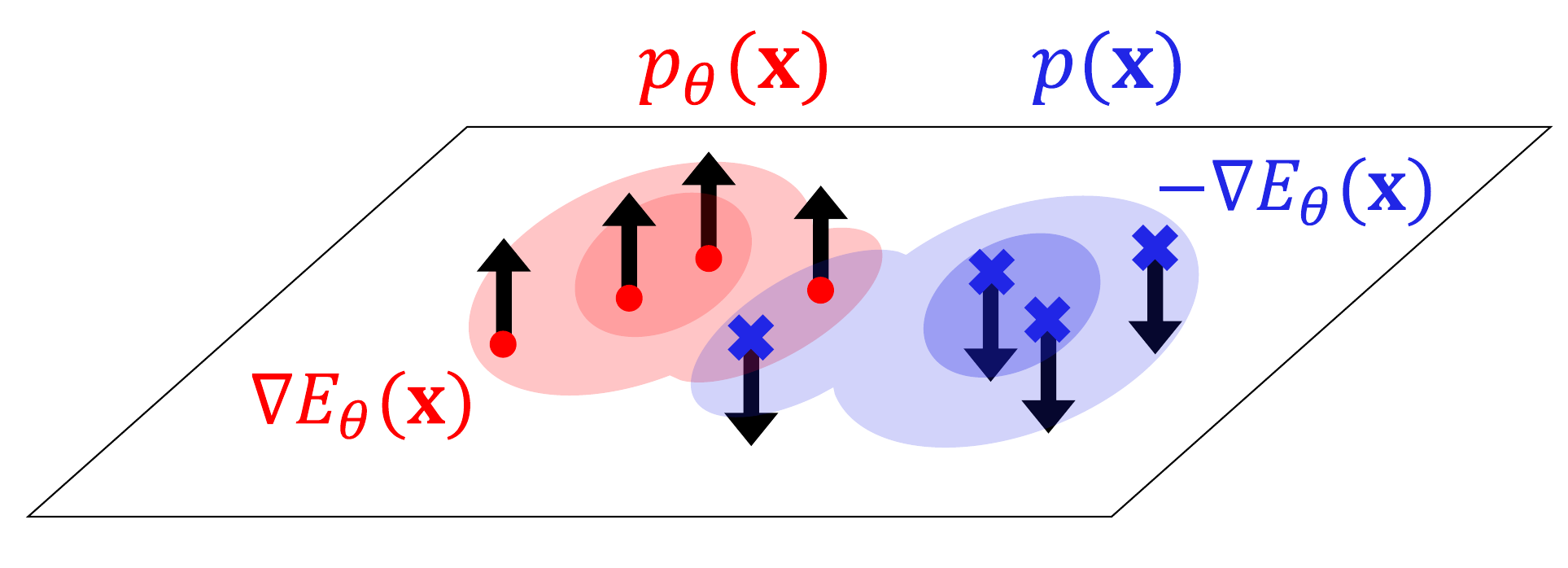}
    \vskip -0.1in
    \caption{An illustration of the energy gradients in Eq.~(\ref{eq:ebm2}). The red and blue shades represent the model and the data density, respectively. The gradient update following Eq.~(\ref{eq:ebm2}) increases the energy of samples from $p_\theta(\bx)$ (the red dots) and decreases the energy of training data (the blue crosses).}
    \label{fig:ebm_training}
    \vskip -0.1in
\end{figure}

In Eq.~(\ref{eq:ebm2}), the first term decreases the energy of the training data, or ``positive" samples, while the second term increases the energy of the generated samples, or ``negative" samples.
The training converges when $p_\theta (\bx)$ becomes identical to $p(\bx)$, as the two gradient terms cancel out. 
In practice, the two expectations in Eq.~(\ref{eq:ebm2}) are approximated with a mini-batch of samples during each iteration.
Figure \ref{fig:ebm_training} visualizes the gradients in Eq.~(\ref{eq:ebm2}).

\textbf{Langevin Monte Carlo (LMC)} The negative samples are generated using MCMC.
LMC (\citet{parisi1981correlation,grenander1994representations}) is a simple yet effective MCMC method used in recent work on deep EBMs \cite{du2019, Grathwohl2020Your,Nijkamp2019nonconvergent}.
In LMC, a starting point $\bx_0$ is drawn from a noise distribution $p_0(\bx)$, typically a Gaussian or uniform distribution.
Starting from $\bx_0$, a Markov chain evolves as follows:
\begin{align} 
\bx_{t+1} = \bx_t + \lambda_\bx\nabla_\bx \log p_\theta(\bx_t) + \sigma_\bx\epsilon_t,
 \label{eq:lmc}
\end{align}
where $\epsilon_t \sim \mathcal{N}(\mathbf{0}, \mathbf{I})$. $\lambda_\bx$ and $\sigma_\bx$ are the step size and the noise parameters, respectively.
A theoretically motivated choice is $2\lambda_\bx = \sigma_\bx^2$, but the parameters are often tweaked separately for better performance \cite{du2019, Grathwohl2020Your,Nijkamp2019nonconvergent}.
As $\nabla_\bx \log p_\theta(\bx)=-\nabla_\bx E(\bx)/T$, tweaking the step size can be seen as adjusting the temperature $T$.

To ensure the convergence of the chain, either Metropolis-Hastings rejection \cite{roberts1996exponential} or annealing of the noise parameter to zero \cite{welling2011bayesian} may be employed, but often omitted in practice.

We discuss specific strategies to evaluate the second term in Eq.~(\ref{eq:ebm2}) in Section \ref{sec:training}.
For a comprehensive review on various strategies for training an EBM, readers may refer to \citet{song2021train}.

\section{Normalized Autoencoders}

\subsection{Definition}

We propose \textbf{Normalized Autoencoder (NAE)}, a normalized probabilistic model defined from an autoencoder.
The probability density of NAE $p_\theta(\bx)$ is defined as a Gibbs distribution (Eq.~(\ref{eq:ebm})) the energy of which is defined as the reconstruction error of an autoencoder:
\begin{align}
    E_\theta(\bx) = l_\theta(\bx). \label{eq:energy_reconstruction_error}
\end{align}
Thus, the model density of NAE is given as
\begin{align}
    p_\theta(\bx) = \frac{1}{\Omega_\theta}\exp(-l_\theta(\bx)/T), \label{eq:nae_def}
\end{align}
where $\Omega_\theta$ is defined as in Eq.~(\ref{eq:Omega}).
Due to the normalization constant, $p_\theta(\bx)$ is a properly normalized probability density.

As a probabilistic model, NAE is trained to maximize the likelihood of data.
The loss function to be minimized is the negative log-likelihood of data:
\begin{align}
    \mathbb{E}_{\bx \sim p(\bx)}[-\log p_\theta(\bx)] = \mathbb{E}_{\bx \sim p(\bx)}[l_\theta(\bx)]/T + \log\Omega_\theta. \label{eq:nae_objective}
\end{align}
The gradient for the negative log-likelihood is evaluated as in conventional EBMs (Eq.~(\ref{eq:ebm2})).
\begin{align}
    &\mathbb{E}_{\bx\sim p(\bx)}[-\nabla_{\theta} \log p_{\theta}(\bx)] \nonumber\\
    &= \mathbb{E}_{\bx\sim p(\bx)}[\nabla_{\theta} l_\theta(\bx)]/T - \mathbb{E}_{\bx'\sim p_{\theta}(\bx)}[\nabla_{\theta} l(\bx')]/T. \label{eq:nae_gradient}
\end{align}
Therefore, each gradient step decreases the reconstruction error of training data $\bx$, while increasing the reconstruction error of negative samples $\bx'$ generated from $p_\theta(\bx)$. 

\subsection{Remarks}

\textbf{Normalization as Regularization} In NAE, enforcement of normalization can be viewed as a regularizer for the reconstruction loss (\ref{eq:ae_objective}).
A typical formulation for a regularized autoencoder is given as $L=L_{\text{AE}} + L_\text{reg}$, where $L_\text{reg}$ is a regularizer.
By setting the loss function of NAE as $L=T\mathbb{E}_{\bx \sim p(\bx)}[-\log p_\theta(\bx)]$, we have $L=L_{\text{AE}} + T\log\Omega_\theta$.
Therefore, the normalization constant contributes as a regularizer: $L_{\text{reg}}=T\log\Omega_\theta$.

\textbf{Suppression of Outlier Reconstruction} During the training of NAE, the reconstruction of an outlier is inhibited by enforcing the normalization constraint.
Given a successful sampling process, the negative samples should cover all high density regions of $p_\theta(\bx)$.
A sample from a high density region of $p_\theta(\bx)$ has a low $l_\theta(\bx)$ by definition (Eq.~(\ref{eq:energy_reconstruction_error})).
Hence, if there exist a reconstructable outlier, which has high $p_\theta(\bx)$ due to low $l_\theta(\bx)$, it will appear as a negative sample from MCMC.
As the gradient update given in Eq.~(\ref{eq:nae_gradient}) increases the reconstruction error of negative samples, the reconstruction quality of a reconstructable outlier will be degraded.
As a result, the reconstruction error of NAE becomes a more informative predictor that discriminates outliers from inliers than that of a conventional autoencoder.

\textbf{Outlier Detection with Likelihood} NAE bridges the two popular outlier detection criteria, namely, the reconstruction error \cite{japkowicz1995novelty} and the likelihood \cite{bishop1994novelty}.
The reconstruction error criterion classifies an input with a large reconstruction error as OOD $l_\theta(\bx) > \tau$, whereas the likelihood criterion predicts an input as an outlier if the log-likelihood is smaller than the threshold $\log p_\theta(\bx) < \tau'$.
These two criteria are equivalent in NAE for appropriately set $\tau$ and $\tau'$, as the reconstruction error and the log-likelihood has a linear relationship: $\log p_\theta(\bx)=-l_\theta(\bx) - \log \Omega_\theta$.
Note that the two criteria rarely coincide in other models, for example, denoising autoencoders (DAE, \citet{vincent2008extracting}), VAE \cite{kingma2013auto}), and DSEBMs \cite{zhao2016energy}, causing confusion on which of the decision rules should be employed for outlier detection.

\textbf{Sample Generation} Samples from $p_\theta(\bx)$ are generated through MCMC.
Unlike VAE, the forward pass of a decoder should not be considered as sample generation.

\section{On-Manifold Initialization}
\label{sec:training}

The main challenge in the training of NAE through Eq.~(\ref{eq:nae_gradient}) is that each iteration requires negative sample generation using MCMC, which is computationally expensive.
In this section, we first discuss the failure modes of popular approximate sampling strategies for EBMs, namely Contrastive Divergence (CD; \citet{hinton2002training}) and Persistent CD (PCD; \citet{tieleman2008training}).
We argue that the method on how the initial state of MCMC is chosen have incurred such failure modes.
Then, we propose on-manifold initialization, an approximate sampling strategy effective in training the NAE.
On-manifold initialization provides a better initial state for MCMC by leveraging the structure of an autoencoder.

There exist other training methods for EBMs which do not rely on MCMC, for example denoising score matching \cite{vincent2011connection} or noise contrastive estimation \cite{gutmann2010noise}, and they may also be applicable to NAE. We leave application of such methods on NAE as future work.

\begin{figure}[t]
  \begin{center}
    \includegraphics[width=0.481\textwidth]{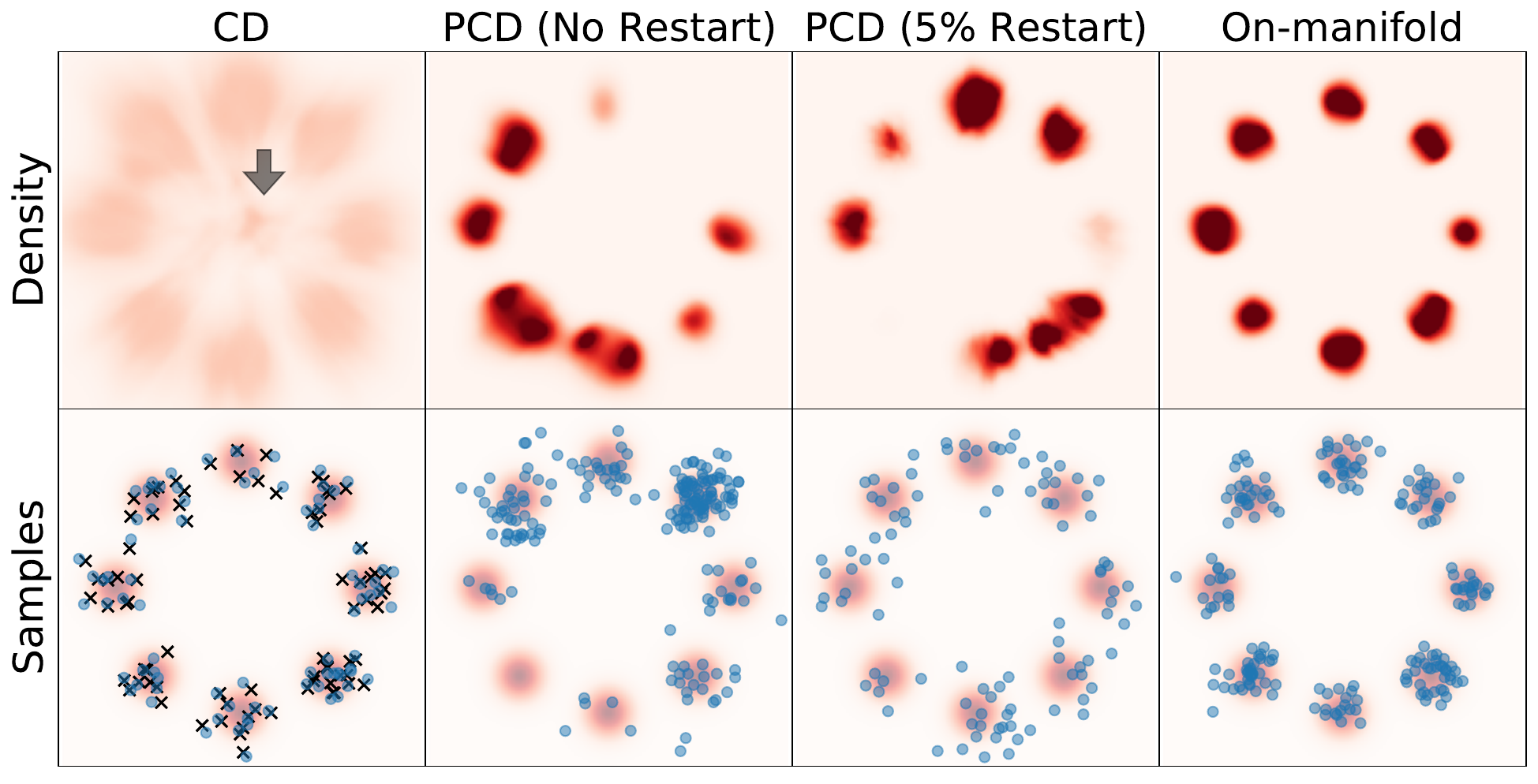}
  \end{center}
  \vskip -0.1in
  \caption{Density estimates and negative samples from NAEs trained by various approximate sampling methods. The generated samples (blue dots) are visualized along with the true density, a 2D mixture of 8 Gaussians. The data density is depicted in Figure \ref{fig:density-estimation}. \textbf{CD}: The learned density has a spurious mode, marked by an arrow. The black crosses denote training data. \textbf{PCD without restart}: The highly correlated samples result in an oscillating density estimate. \textbf{PCD with restart}: Despite the good quality of sampling, the density is poorly estimated. \textbf{On-manifold}: Both density estimation and sample generation are performed well. More details are specified in Section \ref{sec:limitation} and Section \ref{sec:density-estimation}.}
  \label{fig:training_methods}
  \vskip -0.1in
\end{figure}

\subsection{Failure Modes of CD and PCD} \label{sec:limitation}

\textbf{Failure Mode of CD} CD, often called CD-$k$, draws a negative sample by first initializing a Markov chain of MCMC at a training data point, then proceeding $k$ steps of MCMC transitions. 
The strength of CD is that the number of steps $k$ can be radically smaller, e.g., $k=1$, than the usual number of steps required in a convergent MCMC run, significantly reducing the amount of computation.

However, when $k$ is small, CD-$k$ is not able to suppress a spuriously high mode in the model density $p_\theta(\bx)$ located far from the data distribution $p(\bx)$, because negative samples are only generated in the vicinity of training data.
Figure \ref{fig:training_methods} shows an instance of a spurious mode in the model density.
Negative samples (blue dots) are close to training data (black crosses) so that they do not reach for the density mode in the middle.
As a result, the mode is not suppressed.
Such a spurious mode will result in outlier detection failures and, in case of NAE, reconstructed outliers.
The possibility of accidentally assigning high density in the unvisited area was acknowledged in the original article (Section 3 of \citet{hinton2002training}).
Spurious modes are also observed in DAE, where a corrupted datum is located only in the neighborhood of a training data point \cite{alain2014regularized}.
Increasing $k$ will decrease the chance of have spurious modes, but the computational advantage of CD will be lost when $k$ is large.

\textbf{Failure Mode of PCD}
An initial state of MCMC in PCD is given as the negative sample generated from MCMC in the previous training iteration.
PCD was originally implemented using fully persistent MCMC \cite{tieleman2008training}.
However, without a restart, MCMC chains in a mini-batch may become highly correlated to each other.
When $p_\theta(\bx)$ is multi-modal, the correlated chains yield degenerate negative samples which only cover a subset of density modes as in Figure \ref{fig:training_methods}.
The degenerate samples make the density estimate oscillatory, slowing the convergence of the model.

The degeneracy between chains can be mitigated by randomly resetting the initial state to a sample from the noise distribution $p_0(\bx)$ with a small probability (typically 5\%) \cite{du2019,Grathwohl2020Your}.
However, learning with PCD still fails to yield an accurate density estimate (Figure \ref{fig:training_methods}).
This failure mode can be explained by the study of \citet{Nijkamp2019nonconvergent}: When a short MCMC chain initialized from $p_0(\bx)$ is used in training, an EBM simply learns a flow that maps $p_0(\bx)$ to $p(\bx)$, and the energy no longer models the data density.
Using a restart drives an EBM to become such a flow, as restarted chains are short and start from $p_0(\bx)$.

In summary, CD initializes MCMC from the data distribution $p_\theta(\bx)$, and PCD initializes MCMC from a noise distribution $p_0(\bx)$.
The convergence of MCMC is independent of its initialization in theory, but the initialization method can be crucial in practice, as shown in Figure \ref{fig:training_methods}.
When $p_\theta(\bx)$, from which we want to sample, deviates significantly from $p_\theta(\bx)$ or $p_0(\bx)$, these initialization methods may lead to a poor density estimate and a suboptimal performance in outlier detection.

\begin{figure}[t]
    \centering
    \includegraphics[width=0.48\textwidth]{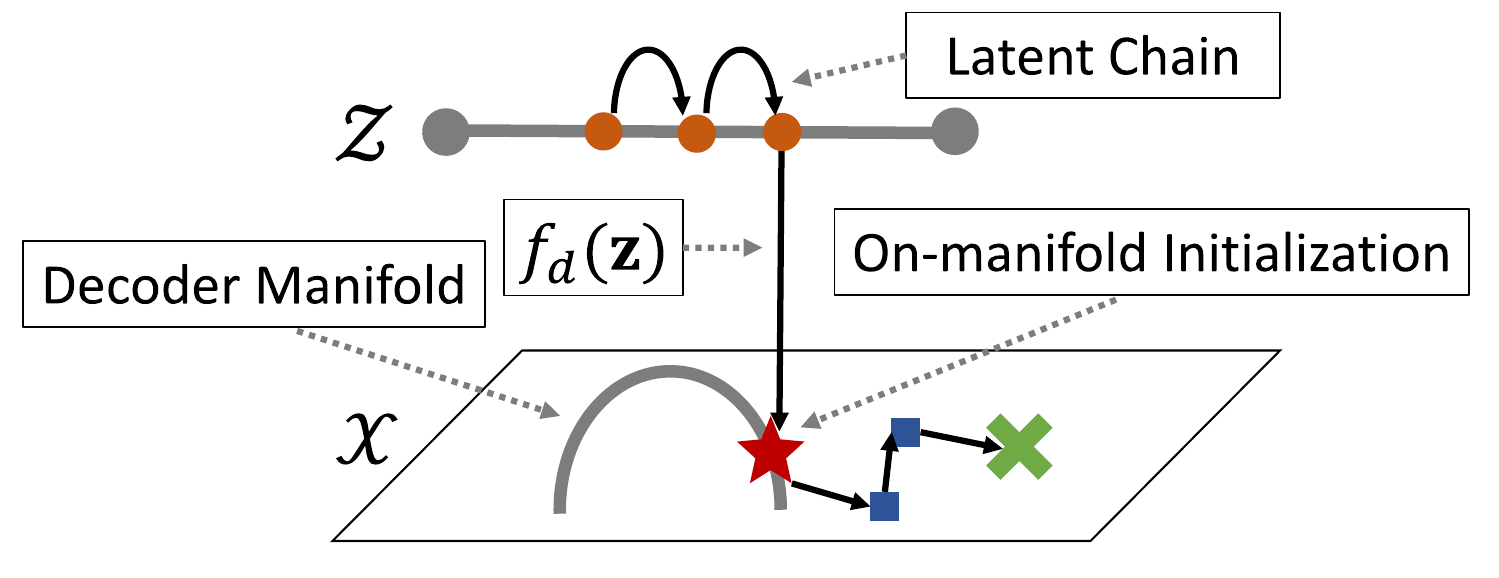}
    \vskip -0.2in
    \caption{An illustration of the on-manifold initialization. The one-dimensional latent space $\mathcal{Z}$ and the two-dimensional input space $\mathcal{X}$ are shown. The red star is the on-manifold initialized state. The cross denotes a negative sample obtained at the end of the whole process.}
    \label{fig:omi}
\end{figure}

\subsection{On-manifold Initialization}

We propose \textbf{on-manifold initialization (OMI)}, a novel MCMC initialization strategy which eventually leads to a significantly better density estimate.
We aim to initialize a MCMC chain from a high-density region of $p_\theta(\bx)$ instead of $p_0(\bx)$ or $p(\bx)$.
While finding a high-density region given an energy function is difficult in general, it is possible for NAE's distribution, since we can exploit the structure of an autoencoder.
For a sufficiently well-trained autoencoder, a point with high $p_\theta(\bx)$, i.e., a small reconstruction error, will lie near the \emph{decoder manifold}, which we define as:
\begin{align}
    \mathcal{M}=\{\bx|\bx=f_d(\bz), \bz \in \mathcal{Z}\}. \label{eq:decoder-manifold}
\end{align}
In on-manifold initialization, we initialize MCMC from a point in the decoder manifold $\bx_0 \in \mathcal{M}$.

Not all points in $\mathcal{M}$ have high $p_\theta(\bx)$.
To find points with high $p_\theta(\bx)$, we run a preliminary MCMC named as \emph{latent chain} in the latent space $\mathcal{Z}$.
The latent chain generates a sample from \emph{on-manifold density} $q_\theta(\bz)$ defined from \emph{on-manifold energy} $H_\theta(\bz)$. 
\begin{align}
    q_\theta(\bz) =& \frac{1}{\Psi_\theta} \exp(-H_\theta(\bz)/T_\bz), \\
    H_\theta(\bz) =& E_\theta(f_d(\bz)),
\end{align}
where $\Psi_\theta = \int \exp(-H_\theta(\bz)/T_\bz) d\bz$ is the normalization constant and $T_\bz$ is the temperature.
A latent vector $\bx$ with a small $H_\theta(\bz)$ will result in a small $E_\theta(\bx)$ when it is mapped to the input space by $\bx=f_d(\bz)$.
Thus, $H_\theta(\bz)$ guides the latent chain to find $\bz$ which produce $\bx_0 \in \mathcal{M}$ which has a small energy, i.e., a small reconstruction error.

Similarly to Eq.~(\ref{eq:lmc}), we use LMC to run the latent chain.
An initial state $\bz_0$ is drawn from a noise distribution defined on the latent space. Then the state propagates as:
\begin{align} 
\bz_{t+1} = \bz_t + \lambda_\bz\nabla_\bz \log q_\theta(\bz_t) + \sigma_\bz\epsilon_t,
 \label{eq:lmc-latent}
\end{align}
where $\lambda_\bz$ and $\sigma_\bz$ are the step size and the noise parameters as in Eq.~(\ref{eq:lmc}).
A sample replay buffer \cite{du2019} is applicable in the latent chain.
Figure \ref{fig:omi} illustrates negative sample generation process using the on-manifold initialization.
We also write the process as an algorithm in Appendix.

\section{Related Work}

\textbf{Autoencoders}
There have been several attempts to formulate a probabilistic model from an autoencoder.
VAE uses a latent variable model by introducing a prior distribution $p(\bz)$. However, the prior may deviate from the actual distribution of data in $\mathcal{Z}$, which may cause problems.
GPND \cite{pidhorskyi2018generative} models probability density by factorizing into on- and off-manifold components but still requires a prior distribution.
$\mathcal{M}$-flow \cite{brehmer2020flows} only defines a probability density on the decoder manifold and does not assign a likelihood to off-manifold data.
DAE models a density by learning the gradient of log-density \citep{alain2014regularized}.

MemAE \cite{gong2019memorizing} is a rare example that directly tackles the outlier reconstruction problem. MemAE employs a memory
module that memorizes training data to prevent outlier
reconstruction, but in this case, the reconstruction error for
an inlier can be large because the model’s generalization ability
is also limited.

\textbf{Design of Energy Functions}
Specifying the class of $E_\theta(\bx)$ not only has computational consequences but alters the inductive bias that an EBM encodes.
Feed-forward convolutional networks are used in \citet{xie2016theory}, \citet{du2019} and \citet{Grathwohl2020Your} and are shown to effectively model the distribution of images.
The energy can also be modeled in an auto-regressive manner \cite{nash2019autoregressive,meng2020autoregressive}. Auto-regressive energy functions are very flexible and thus are capable of modeling high-frequency patterns in data.
VAEBM \cite{xiao2021vaebm} combines VAE and a feed-forward EBM to model complicated data distribution.

The reconstruction error of an autoencoder is used as a discriminator in EBGAN \cite{zhao2016energy}. Although the reconstruction error was called ``energy" in EBGAN, the formulation is clearly different from NAE.
EBGAN does not utilize Gibbs distribution formulation (Eq.~(\ref{eq:ebm})) to model a distribution, and samples are generated from a separate generator network.
In DSEBM \cite{zhao2016energy}, the difference between an input and its reconstruction is interpreted as the gradient of log-density.

\begin{table*}
\footnotesize
\setlength\tabcolsep{1.5pt} 
  \caption{MNIST hold-out class detection AUC scores. The values in parentheses denote the standard error of mean after 10 training runs.}
  \label{tb:mnist-holdout}
  \centering
  \begin{sc}
  \begin{tabular}{llllllllllll}
    \toprule
    Hold-out: & 0     & 1 &2 &3 &4 &5 &6 &7 &8 &9 & avg \\
    \midrule
NAE-OMI & \textbf{.989}{\tiny(.002)} & \textbf{.919}{\tiny(.013)} & \textbf{.992}{\tiny(.001)} & \textbf{.949}{\tiny(.004)} & \textbf{.949}{\tiny(.005)} & \textbf{.978}{\tiny(.003)} & \textbf{.938}{\tiny(.004)} & \textbf{.975}{\tiny(.024)} & \textbf{.929}{\tiny(.004)} & \textbf{.934}{\tiny(.005)} & \textbf{.955}\\
NAE-CD & .799 & .098 & .878 & .769 & .656 & .806 & .874 & .537 & .876 & .500 & .679 \\
NAE-PCD & .745 & .114 & .879 & .754 & .690 & .813 & .872 & .509 & .902 & .544 & .682 \\
AE & .819 & .131 & .843 & .734 & .661 & .755 & .844 & .542 & .902 & .537 & .677\\
DAE & .769 & .124 & .872 & .935 & .884 & .793 & .865 & .533 & .910 & .625 & .731\\
VAE(R) & .954 & .391 & .978 & .910 & .860 & .939 & .916 & .774 & .946 & .721 & .839\\
VAE(L) & .967 & .326 & .976 & .906 & .798 & .927 & .928 & .751 & .935 & .614 & .813\\
WAE & .817 & .145 & .975 & \textbf{.950} & .751 & .942 & .853 & .912 & .907 & .799 & .805\\
GLOW & .803 & .014 & .624 & .625 & .364 & .561 & .583 & .326 & .721 & .426 & .505\\
PXCNN++ & .757 & .030 & .663 & .663 & .483 & .642 & .596 & .307 & .810 & .497 & .545\\
IGEBM & .926 & .401 & .642 & .644 & .664 & .752 & .851 & .572 & .747 & .522 & .672\\
DAGMM & .386 & .304 & .407 & .435 & .444 & .429 & .446 & .349 & .609 & .420 & .423\\
    \bottomrule
  \end{tabular}
  \end{sc}
  \vspace{-0.5cm}
\end{table*}

\section{Experiments}

\subsection{Technicalities for NAE Training}

\textbf{Pre-training as a Conventional Autoencoder}
NAE can be pre-trained as a conventional autoencoder by minimizing the reconstruction error following Eq.~(\ref{eq:ae_gradient}), before the main training.
By providing a good initialization for network weights and the decoder manifold, pre-training greatly reduces the number of NAE training iterations (Eq.~(\ref{eq:nae_gradient})) required until convergence.
Pre-training is not always necessary: In our experiments, we observe that NAE can be trained successfully without pre-training for synthetic data.
However, pre-training was essential to obtain decent results for larger scale data, such as MNIST and CIFAR-10.

\textbf{Latent Space Structure}
Two configurations for the latent space is used in experiments: the unbounded real space $\RE^{D_\bz}$ and the surface of a hypersphere $\mathbb{S}^{D_\bz-1}$.
When $\mathcal{Z}=\RE^{D_\bz}$, a linear layer is used as the output of an encoder.
$q_0(\bz)$ is set as $\mathcal{N}(\mathbf{0},\mathbf{I})$.
The squared norm of the latent vectors are added to the loss function as a regularizer so that $\bz$'s concentrate near the origin \cite{Ghosh2020From}.

For the hyperspherical space $\mathcal{Z}=\mathbb{S}^{D_\bz -1}$ \cite{davidson2018hyperspherical,xu2018spherical,zhao2019latent}, the output of an encoder is projected to the surface of a unit ball through the division by its norm: $\bz \leftarrow \bz / ||\bz ||$.
In Langevin dynamics, a sample is projected to $\mathbb{S}^{D_\bz -1}$ at the end of each step.
$q_0(\bz)$ is set to a uniform distribution on $\mathbb{S}^{D_\bz -1}$.

The hyperspherical latent space has a few advantages over $\RE^{D_{\bz}}$.
First, it is impossible to draw uniformly random samples, because $\RE^{D_\bz}$ is not compact. 
Second, for large $D_\bz$, it is difficult to draw samples near the origin, because of its exponentially decreasing volume.
However, we believe more works needs to be done to completely understand the effect of hyperspherical geometry on the latent representation.

\textbf{Regularizing Negative Sample Energy}
As introduced in \citet{du2019}, we regularize the energy of negative samples to prevent its divergence.
We add the average squared energy of negative samples in a mini-batch to the loss function: $L = L_\text{NAE} + \alpha \sum_{i=1}^{B} E(\bx'_i)^2/B$ for the batch size $B$ and the hyperparameter $\alpha$. We set $\alpha=1$.

\begin{figure}
  \begin{center}
    \includegraphics[width=0.48\textwidth]{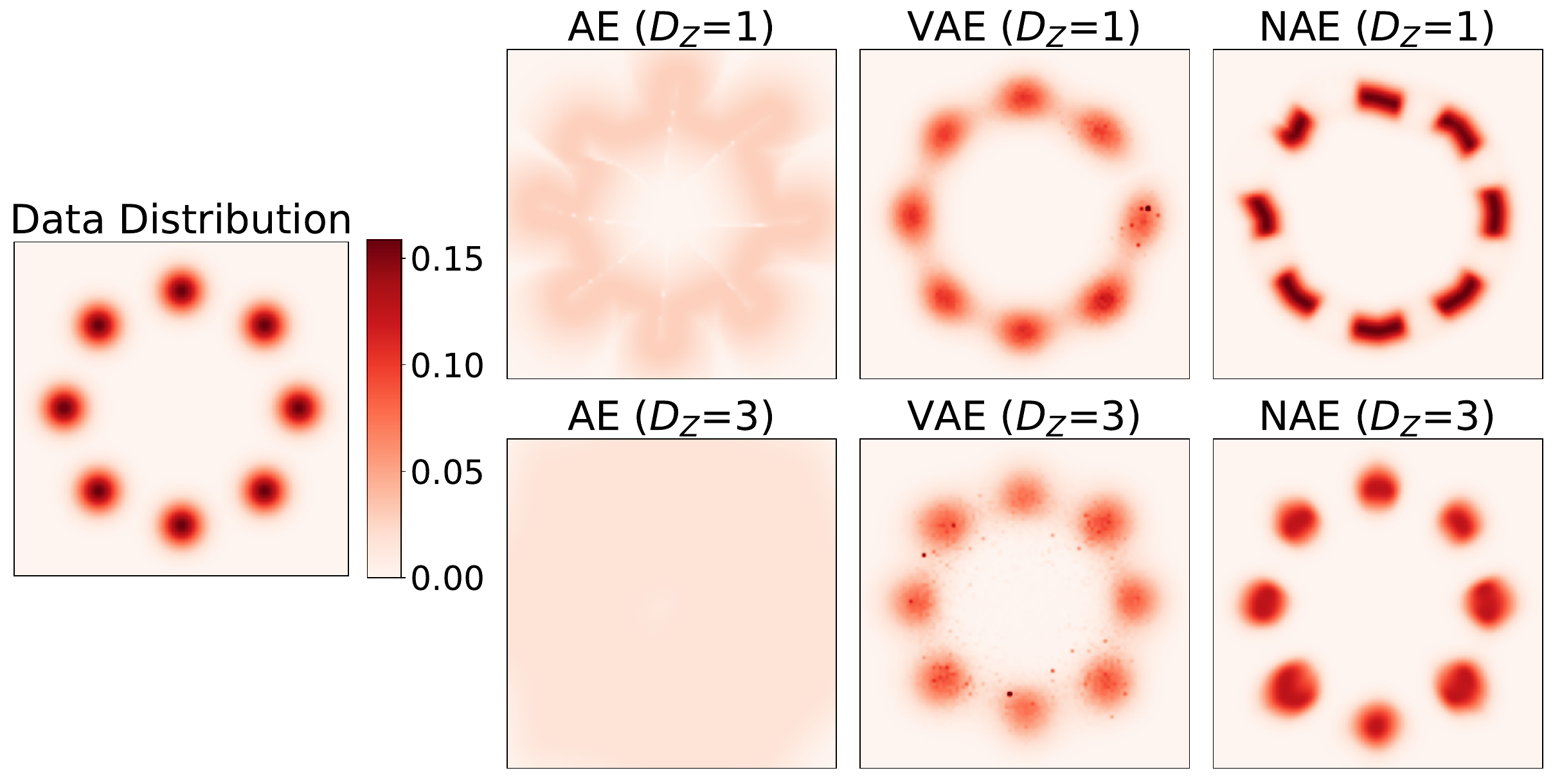}
  \end{center}
  \vskip -0.1in
  \caption{Estimating 8 Gaussians using various autoencoders. The density of an autoencoder (AE) is computed from Eq.~(\ref{eq:nae_def}). AE gives a significant amount of probability to low-data-density area.
  VAE also assigns some probability mass in between Gaussians.
  Meanwhile, the density estimate from NAE agrees well with the data distribution.}
  \label{fig:density-estimation}
  \vskip -0.2in
\end{figure}

\subsection{2D Density Estimation} \label{sec:density-estimation}

We demonstrate the density estimation capability of NAE with a two-dimensional mixture of 8 Gaussians. First, we benchmark negative sample generation strategies for NAE, including CD, PCD with and without restart, and on-manifold initialization.
The results are shown in Figure \ref{fig:training_methods} and discussed in Section \ref{sec:limitation} in detail.

Second, we compare NAE trained with the on-manifold initialization to a conventional autoencoder and VAE (Figure \ref{fig:density-estimation}).
An autoencoder assigns high densities on regions between Gaussian modes, meaning that an autoencoder gives a small reconstruction error from a points from the region.
For the overcomplete case ($D_\bz=3 > D_\bx$), an autoencoder almost becomes the identity map, and its reconstruction error is not an informative predictor for an outlier.
VAE and NAE learn a non-identity function under the overcomplete setting, showing the effectiveness of their regularizers.

In the experiments, the identical network architecture is used, and the temperature is optimized by gradient descent.
In on-manifold initialization, temperature values are shared by the main MCMC and the latent chain.
When performing MCMC in $\mathcal{X}$, Metropolis-Hastings rejection is applied to ensure the detailed balance but is not applied in the latent chain.
For visualization, the normalization constants for an autoencoder and NAE are computed by numerically integrating over the domain, $[-4,4]^2$.

\subsection{Outlier Detection} \label{sec:exp-outlier-detection}

\textbf{Experimental Setting}
We empirically demonstrate the effectiveness of NAE as an outlier detector.
In outlier detection tasks, an outlier detector is trained only using inlier data and then asked to discriminate outliers from inliers during test phase.
Given an input, a detector is assumed to produce a scalar decision function which indicates the outlierness of the input. 
We measure the detection performance in AUC, i.e., the area under the receiver operating characteristic curve.
Following the protocol of \citet{ren2019likelihood} and \citet{hendrycks2018deep}, we use an OOD dataset different from the datasets used in test phase to tune model hyperpamraeters. Additional details on model implementation and datasets can be found in the supplementary material.

The identical networks architectures are used for all autoencoder-based methods.
The reconstruction error is used as the decision function, except for VAE.
For deep generative models, PixelCNN++ (PXCNN++, \citet{salimans2017pixelcnn++}), Glow \cite{kingma2018glow} and a feed-forward EBM (IGEBM, \citet{du2019}), we use the negative log-likelihood (i.e., the energy) as the decision function.
For VAE, we show two results from using the reconstruction error (R) or the negative log-likelihood (L) as decision functions.

\textbf{MNIST Hold-Out Class Detection}
One class from MNIST is set as the outlier class and the rest as the inlier class. Then, the procedure is repeated for all ten classes in MNIST.
ConstantGray dataset is used for model selection.

\begin{figure}
    \centering
    \includegraphics[width=0.45\textwidth]{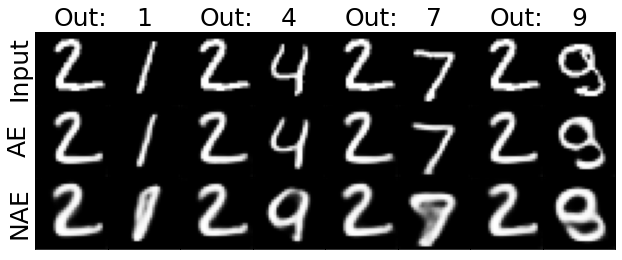}
    \vskip -0.1in
    \caption{Reconstruction examples in MNIST hold-out class detection. Data and their reconstructions are shown for four difficult hold-out settings (1, 4, 7 and 9). Digit 2 is shown as an inlier example. The bottom two rows depict the reconstructions from four autoencoders (AE) and four NAEs trained on each setting. AEs reconstruct the outlier class well, while NAEs selectively reconstruct only inliers.}
    \label{fig:mnist-ho-recon}
    \vskip -0.2in
\end{figure}

This problem is not as easy as it seems, as confirmed in the very low performance of various algorithms in Table \ref{tb:mnist-holdout}.
When a class is held out from MNIST, the remaining 9 classes may contain a set of visual features sufficient to reconstruct the hold-out class, i.e., the outlier reconstruction occurs.
The outlier reconstruction is particularly severe for the digit 1, 4, 7 and 9, possibly because their shape can be reconstructed from the recombination of other digits. For example, overlapping 4 and 7 produces a shape similar to 9.
Interestingly, most of the other baseline algorithms also show poor performance when 1, 4, 7 or 9 are held out as the outlier.
NAE shows the highest AUC score for all classes and effectively suppresses the reconstruction of the outlier class (Figure \ref{fig:mnist-ho-recon}).

We also compare CD and PCD along with OMI in training NAEs.
Using CD and PCD show poor outlier detection performance, although given the identical set of MCMC parameters.

\begin{table}[t]
  \vspace{-10pt}
\footnotesize
\setlength\tabcolsep{1.5pt} 
  \caption{OOD detection performance in AUC.}
  \label{tb:cifar-ood}
  \centering
  \begin{tabular}{cccccc}
    \toprule
    In: CIFAR-10 & ConstantGray     & FMNIST & SVHN & CelebA  & Noise  \\
    \midrule
NAE & \textbf{.963} & \textbf{.819} & \textbf{.920} & \textbf{.887} & 1.0\\
AE & .006 & .650 & .175 & .655 & 1.0\\
DAE & .001 & .671 & .175 & .669 & 1.0\\
VAE(R) & .002 & .700 & .191 & .662 & 1.0\\
VAE(L) & .002 & .767 & .185 & .684 & 1.0\\
WAE & .000 & .649 & .168 & .652 & 1.0\\
GLOW & .384 & .222 & .260 & .419 & 1.0\\
PXCNN++ & .000 & .013 & .074 & .639 & 1.0\\
IGEBM & .192 & .216 & .371 & .477 & 1.0\\
    \bottomrule
  \end{tabular}

\footnotesize
\setlength\tabcolsep{1.0pt} 
  \centering
  \begin{tabular}{cccccc}
    \toprule
    In: ImageNet32 & ConstantGray     & FMNIST & SVHN & CelebA  & Noise  \\
    \midrule
NAE & \textbf{.966} & \textbf{.994} & \textbf{.985} & \textbf{.949} & 1.0\\
AE & .005 & .915 & .102 & .325 & 1.0\\
DAE & .069 & .991 & .102 & .426 & 1.0\\
VAE(R) & .030 & .936 & .132 & .501 & 1.0\\
VAE(L) & .028 & .950 & .132 & .545 & 1.0\\
WAE & .069 & .991 & .081 & .364 & 1.0\\
GLOW & .413 & .856 & .169 & .479 & 1.0\\
PXCNN++ & .000 & .004 & .027 & .238 & 1.0\\
    \bottomrule
  \end{tabular}
  \vskip -0.1 in
\end{table}

\textbf{Out-of-Distribution Detection}
The samples from different datasets are used as the outlier class.
We test two inlier datasets, CIFAR-10 or ImageNet 32$\times$32 (ImageNet32).
Zero-padded 32$\times$32 MNIST images are used for model selection.
Results are shown in Table \ref{tb:cifar-ood}. 

It is known that constant images and SVHN images are particularly difficult outliers for generative models trained on a set of images with rich visual features \cite{nalisnick2018do,Serra2020Input}.
However, NAE detect such difficult outliers successfully.
All models are able to discriminate noise outliers, indicating that their poor performance is not from the failure of training.

\subsection{Sample Generation}
\begin{figure}[t]
    \centering
    \includegraphics[width=0.48\textwidth]{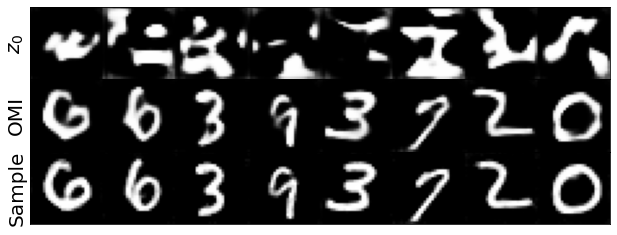}
    \vskip - 0.05 in
    \includegraphics[width=0.48\textwidth]{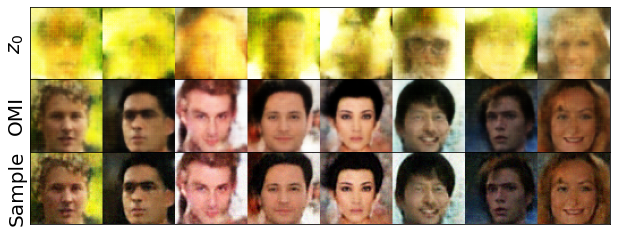}
    \vskip - 0.1 in
    \caption{Sampling with NAEs trained on MNIST and CelebA 64$\times$64. ($\bz_0$) The random initialization of the latent chain. We visualize $f_d(\bz_0)$. (OMI) Images after OMI. (Samples) Samples obtained after MCMC starting from OMI. OMI images and Samples corresponds to the red start and the green cross in Figure \ref{fig:omi}, respectively.}
    \label{fig:sample-mnist-celeba}
    \vskip - 0.1 in
\end{figure}

Samples are generated from NAE using MCMC with OMI.
Figure \ref{fig:sample-mnist-celeba} shows the samples from NAEs trained on MNIST and on CelebA 64$\times$64. The random initial states of the latent chain ($\bz_0$) map to unrecognizable images. After the latent chain, OMI produces somewhat realistic images.
MCMC on $\mathcal{X}$ refines the OMI images.
Although quantitative image (in Appendix) quality metric for samples generated from NAE is not on a par with that of generative models which specialize in sampling, but the generated samples are indeed visually sensible.

\section{Discussion and Conclusion}

\textbf{Comparison to Other EBMs}
NAE uses Gibbs distribution to define a density function as in other EBMs (Eq.~\ref{eq:ebm}).
The main difference between NAE and other EBMs is the choice of an energy function.
However, this difference results in significant theoretical and practical consequences.
First, we naturally incorporate the manifold hypothesis, i.e. the assumption that high-dimensional data lie on a low-dimensional manifold, into a model.
Second, the energy function of NAE can be pre-trained as a conventional autoencoder. 
Third, more effective sampling can be performed by using OMI, leading to a more accurate density estimate.

\textbf{Likelihood-based Outlier Detection and Inductive Bias}
The likelihood is considered as a poor decision function for outlier detection, after the failures of likelihood-based deep generative models such as VAE, PixelCNN++, and Glow \cite{nalisnick2018do,hendrycks2018deep}.
Those generative models fail to detect obvious outlier images which typically have low complexity.
However, we believe that the failures should not be attributed to the use of the likelihood.
There are likelihood-based models, particularly EBMs \cite{du2019,Grathwohl2020Your}, including NAE, that show better outlier detection performance than VAE, PixelCNN++ and Glow.
Instead, inductive bias of a generative model is likely to be responsible for the failure of detecting low-complexity outliers.
It is reported that the likelihoods of the failed models are negatively correlated to the complexity of images \cite{Serra2020Input}.
Meanwhile, the reconstruction of low-complexity images are explicitly suppressed in NAE training, as the simple images tend to lie on the decoder manifold.


\begin{figure}
  \begin{center}
    \includegraphics[width=0.41\textwidth]{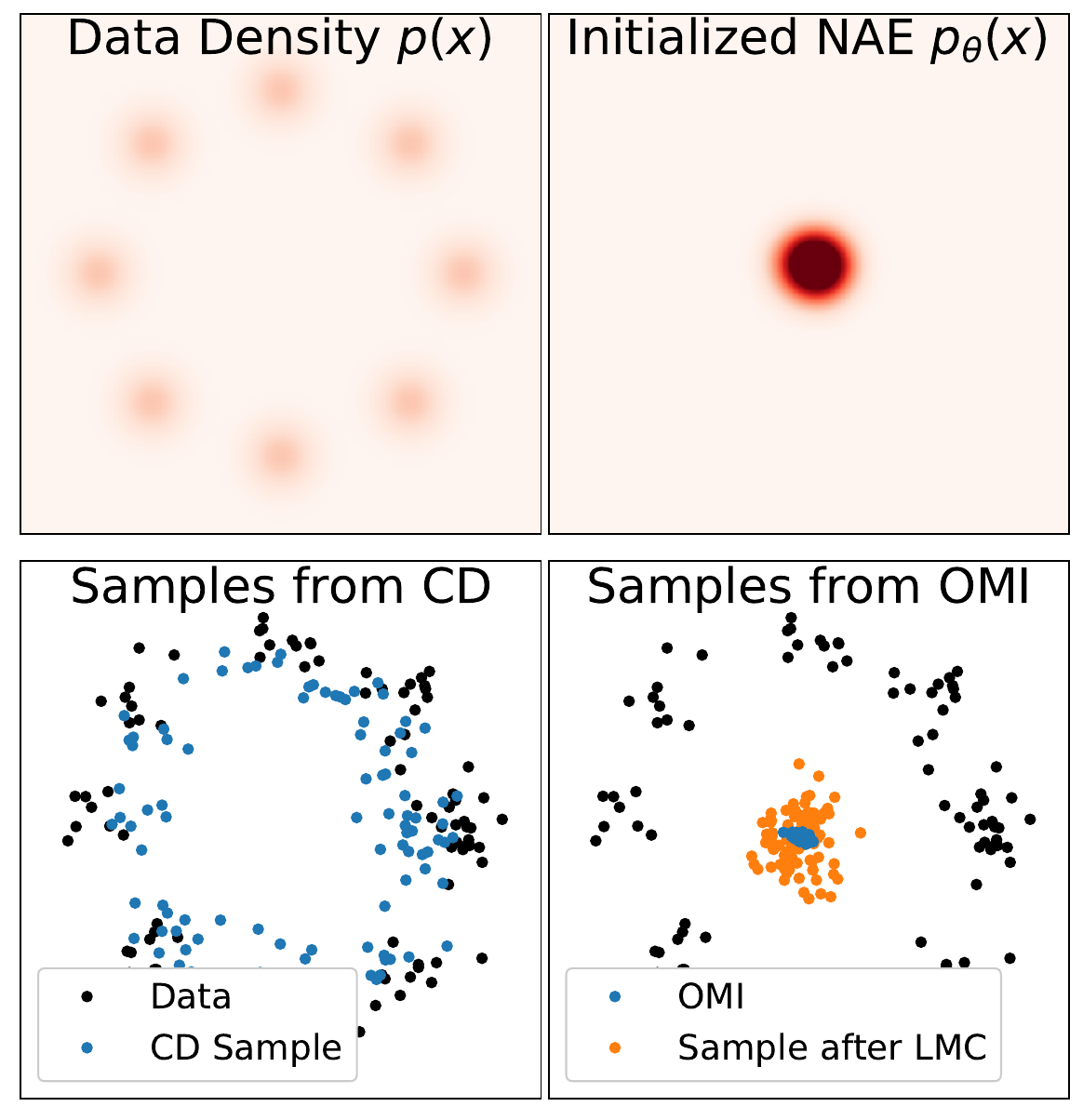}
  \end{center}
  \vspace{-10pt}
  \caption{Sampling with randomly initialized NAE.}
  \label{fig:early}
  \vspace{-10pt}
\end{figure}

\textbf{OMI in Early Stage of Training}
Sampling with OMI generates samples with high model density $p_\theta(\bx)$ even \emph{in the early stage of training}.
In fact, the early stage is where the advantage of OMI over CD is salient, because $p_\theta(\bx)$ differs from $p(\bx)$ significantly.
Figure \ref{fig:early} visualizes samples generated via CD and OMI from a randomly initialized NAE and shows that CD fails to draw samples from $p_\theta(\bx)$.

OMI draws high-model-density proposals because it is designed to exploit the assumption that well-reconstructed points lie on the decoder manifold.
We find that this assumption holds well for all experimental settings used in the paper.

\textbf{Analytic Solution for Linear Case}
Linear NAEs reduce to Gaussian distributions.
Consider $f_e(\bx)=W\bx$ and $f_d(\bz)=W^\top \bz$ with $W \in \RE^{D_\bz \times D_\bx}$. 
Given the squared $L^2$ distance reconstruction error, the density of NAE is written as:
\begin{align}
    p_\theta(\bx) = \exp(-\bx^\top \Sigma^{-1}\bx/2) / \Omega_\theta,
\end{align}
where $\Sigma^{-1}=2(I-W^\top W)^2/T$.
When the determinant of $I-W^\top W$ is non-zero, $p_\theta(\bx)$ becomes a well-defined Gaussian.
Under certain conditions (see Appendix), the maximum likelihood estimate of $\Sigma$ becomes the empirical covariance of data, as in a usual Gaussian distribution.

It is interesting to note that a linear VAE also reduces into a Gaussian, as it is equivalent to probabilistic PCA\cite{kingma2013auto}.
On the other hand, a linear autoencoder is equivalent to PCA \cite{bourlard1988auto}, which is not a generative model.

\textbf{Conclusion}
We have introduced a novel interpretation of the reconstruction error as an energy function.
Our interpretation leads to a novel class of probabilistic autoencoders, which shows impressive OOD detection performance and bridges EBMs and autoencoders.

\section*{Acknowledgements}

S. Yoon and F.C. Park were supported in part by the SNU Artificial Intelligence Institute (AIIS), NRF Grant 2016R1A5A1938472 (SRRC), the SNU BK21+ Program in Mechanical Engineering, and SNU-IAMD.
Y.-K. Noh was partly supported by NRF/MSIT (No. 2018R1A5A7059549, 2021M3E5D2A01019545),  IITP/MSIT Artificial Intelligence Graduate School Program for Hanyang University (2020-0-01373), and Hanyang University (HY-2019).

\bibliography{ref}
\bibliographystyle{icml2021}

\clearpage
\appendix

The Appendix is organized as follows:
Section A provides mathematical derivations, Section B contains extended experimental results,
Section C addresses additional topics for discussions,
and experimental details are provided in Section D.

\section{Derivations}
\subsection{Derivation for Log-Likelihood Gradient in EBM} \label{sec:ebm-grad-deriv}

Here, we present the derivation for the gradient of log likelihood in Eq.~(\ref{eq:ebm2}).
\begin{align*}
    &\mathbb{E}_{\bx\sim p(\bx)}[\nabla_{\theta} \log p_{\theta}(\bx)] \\
    &= - \mathbb{E}_{\bx\sim p(\bx)}[\nabla_{\theta} E(\bx)]/T + \mathbb{E}_{\bx'\sim p_{\theta}(\bx)}[\nabla_{\theta} E(\bx')]/T.
\end{align*}
This expression is well-known in EBM literature \cite{younes1999convergence}, but we provide its derivation to make the paper self-contained.

Recall that the model density function $p_{\theta}(\bx)$ is defined from an energy function $E_{\theta}(\bx)$ using Gibbs distribution: $p_{\theta}(\bx) = \exp(-E_\theta(\bx)/T)/\Omega_\theta$ for the normalization constant $\Omega_\theta = \int \exp(-E_\theta(\bx)/T) \mathrm{d}\bx < \infty$ with the temperature $T$. The gradient for the log likelihood of a single datum $\bx$ is given as follows: 
\begin{align*}
    \nabla_{\theta}&\log p_\theta (\bx) \nonumber \\
    =& - \nabla_{\theta} E(\bx)/T-\nabla_{\theta}\log\Omega_{\theta} \\
    =& -\nabla_{\theta} E(\bx)/T- \frac{\nabla_{\theta} \Omega_\theta}{\Omega_\theta} \\
    =& -\nabla_{\theta} E(\bx)/T- \frac{1}{\Omega_\theta} \nabla_{\theta} \int \exp(-E_\theta(\bx')/T) \mathrm{d}\bx' \\
    =& -\nabla_{\theta} E(\bx)/T- \frac{1}{\Omega_\theta}  \int \nabla_{\theta} \exp(-E_\theta(\bx')/T) \mathrm{d}\bx' \\
    =& -\nabla_{\theta} E(\bx)/T \\
    &\;\;+  \int \frac{1}{\Omega_\theta} \exp(-E_\theta(\bx')/T) \nabla_{\theta} E_\theta(\bx')/T  \mathrm{d}\bx' \\
    =& -\nabla_{\theta} E(\bx)/T + \mathbb{E}_{\bx'\sim p_\theta(\bx')}\left[ \nabla_{\theta} E_{\theta}(\bx')/T\right].
\end{align*}

Taking  the expectation over the data density $p(\bx)$,
\begin{align*}
    &\mathbb{E}_{\bx\sim p(\bx)}[ \nabla_{\theta}\log p_\theta (\bx)] \\
    &=
    -\mathbb{E}_{\bx\sim p(\bx)}\left[\nabla_{\theta} E(\bx)/T\right]\\
    &\;\;\;\;+ \mathbb{E}_{\bx\sim p(\bx)}\left[\mathbb{E}_{\bx'\sim p_\theta(\bx')}\left[ \nabla_{\theta} E_{\theta}(\bx')/T\right]\right] \\
    &= -\mathbb{E}_{\bx\sim p(\bx)}\left[\nabla_{\theta} E(\bx)/T\right] + \mathbb{E}_{\bx'\sim p_\theta(\bx')}\left[ \nabla_{\theta} E_{\theta}(\bx')/T\right].
\end{align*}
Thus, we obtain Eq.~(\ref{eq:ebm2}) of the main manuscript.

\subsection{Analytic Solution for Linear Case}
We provide a more detailed derivation for linear NAE.
When a linear overcomplete autoencoder is used, NAE reduces into a Gaussian distribution.
Consider a linear deterministic encoder $f_e(\bx)=W\bx$ and a decoder $f_d(\bz)=W^\top \bz$, where $W\in\RE^{D_\bz \times D_\bx}$.
For the squared $L^2$ distance reconstruction error,
\begin{align}
    l(\bx)&=||\bx - W^\top W \bx||^2\\
    &=\bx^\top (I-W^\top W)^2\bx,
\end{align}
and therefore the density of NAE can be written as:
\begin{align}
    p_\theta(\bx) = \frac{1}{\Omega_\theta}\exp(-\bx^\top \Sigma^{-1}\bx/2), \label{eq:nae-linear}
\end{align}
where $\Sigma^{-1} = 2(I-W^\top W)^2/T$. 

Eq.~(\ref{eq:nae-linear}) is a Gaussian distribution with zero mean and $\Sigma$ covariance.
For this Gaussian to be well-defined, the normalization constant should be finite $\Omega_\theta<\infty$.
To the covariance positive definite, we need that the determinant of $(I-W^\top W)$ is non-zero, i.e., no eigenvalue of $W^\top W$ should be one. This means that the autoencoder should not be the identity along any of orthogonal bases.

As an interesting special case, consider $W=0$.
This zero-autoencoder is uninformative, since all inputs are mapped to the origin, but it still defines a valid probability distribution.
In fact, $p_\theta(\bx)$ becomes a standard normal distribution.

Now, we consider the overcomplete setting, where $D_\bz \geq D_\bx$, and look for the maximum likelihood parameter estimate. 
When $D_\bz \geq D_\bx$, the matrix $(I-W^\top W)^2$ spans all positive semidefinite matrices.
Given a zero-centered dataset $\mathcal{D}=\{\bx_{i}\}_{i=1}^{N}$, $\Sigma$ that maximizes the the likelihood of data is the empirical covariance $\Sigma_{\text{ML}}=\sum_{i=1}^{N}\bx_i\bx_i^\top / N$.
Therefore, NAE trained to maximum the likelihood of data is identical to a Gaussian distribution fitted via maximum likelihood.

Recall that a conventional linear autoencoder becomes the identity when $D_\bz \geq D_\bx$.

\begin{table*}
\footnotesize
\setlength\tabcolsep{1.5pt} 
  \caption{MNIST hold-out class detection AUC scores. The values in parentheses denote the standard error of mean after 10 training runs. The values below the horizontal divider line are the newly appended values.}
  \label{tb:mnist-holdout-ext}
  \centering
  \begin{sc}
  \begin{tabular}{llllllllllll}
    \toprule
    Hold-out: & 0     & 1 &2 &3 &4 &5 &6 &7 &8 &9 & avg \\
    \midrule
NAE-OMI & \textbf{.989}{\tiny(.002)} & \textbf{.919}{\tiny(.013)} & \textbf{.992}{\tiny(.001)} & \textbf{.949}{\tiny(.004)} & \textbf{.949}{\tiny(.005)} & \textbf{.978}{\tiny(.003)} & \textbf{.938}{\tiny(.004)} & \textbf{.975}{\tiny(.024)} & \textbf{.929}{\tiny(.004)} & \textbf{.934}{\tiny(.005)} & \textbf{.955}\\
NAE-CD & .799 & .098 & .878 & .769 & .656 & .806 & .874 & .537 & .876 & .500 & .679 \\
NAE-PCD & .745 & .114 & .879 & .754 & .690 & .813 & .872 & .509 & .902 & .544 & .682 \\
AE & .819 & .131 & .843 & .734 & .661 & .755 & .844 & .542 & .902 & .537 & .677\\
DAE & .769 & .124 & .872 & .935 & .884 & .793 & .865 & .533 & .910 & .625 & .731\\
VAE(R) & .954 & .391 & .978 & .910 & .860 & .939 & .916 & .774 & .946 & .721 & .839\\
VAE(L) & .967 & .326 & .976 & .906 & .798 & .927 & .928 & .751 & .935 & .614 & .813\\
WAE & .817 & .145 & .975 & \textbf{.950} & .751 & .942 & .853 & .912 & .907 & .799 & .805\\
GLOW & .803 & .014 & .624 & .625 & .364 & .561 & .583 & .326 & .721 & .426 & .505\\
PXCNN++ & .757 & .030 & .663 & .663 & .483 & .642 & .596 & .307 & .810 & .497 & .545\\
IGEBM & .926 & .401 & .642 & .644 & .664 & .752 & .851 & .572 & .747 & .522 & .672\\
DAGMM & .386 & .304 & .407 & .435 & .444 & .429 & .446 & .349 & .609 & .420 & .423\\
\midrule
VQVAE & .937 &	.272 &	.915 &	.807 &	.673 &	.807 &	.892 &	.643 &	.816 &	.596 & .736 \\
Ganomaly(Repro) & .418 &	.676 &	.479 &	.750 &	.591 &	.615 &	.480 &	.546 &	.551 &	.427 & .553 \\
Ganomaly(Pub)& .740 & .142 & .873 & .699 & .695 & .725 & .787 & .502 & .840 & .508 & .651 \\
    \bottomrule
  \end{tabular}
  \end{sc}
  \vspace{-0.5cm}
\end{table*}

\section{Extended Experimental Results}

In this section, we provide additional experimental results.

\subsection{MNIST Hold-Out Class Detection}

Table \ref{tb:mnist-holdout-ext} augments Table \ref{tb:mnist-holdout} by providing results from VQVAE \cite{oord2017neural} and Ganomaly \cite{akcay2018ganomaly}.

We could not experiment with MemAE \cite{gong2019memorizing} as the training code was not provided by the authors. Instead, we experiment with VQVAE, since it has a similar inductive bias that the only finite number of discrete latent representation can be used.

For Ganomaly, we report two versions of results. Ganomaly(Repro) indicates our re-implementation that shares the network architectures with other autoencoders used in the experiment.
Ganomaly(Pub) is the result published by the original paper. We obtained the numbers by running the scripts provided by the authors' official repository.

As shown in Table \ref{tb:mnist-holdout-ext}, VQVAE and Ganomaly show unsatisfactory performance in detecting the hold-out digit.

\subsection{Out-of-Distribution Detection}

We repeat OOD detection experiment in Section \ref{sec:exp-outlier-detection} with MNIST as the in-distribution dataset. The results are present in Table \ref{tb:MNIST-OOD}.

We use images from Constant Gray, HalfMNIST, ChimeraMNIST, Omniglot, FashionMNIST and Noise dataset for OOD inputs.
The whole MNIST training set without holding out is used to train models.
As shown in Figure \ref{fig:mnist_recon}, an autoencoder shows limited performance due to outlier reconstruction. 
Note that the blank image used in Figure \ref{fig:mnist_recon} is an instance of ConstantGray dataset.
However, NAE successfully detects all the specified OOD inputs.
We omit FashionMNIST results from Table \ref{tb:MNIST-OOD}, because all models detects FashionMNIST inputs with AUC very close to 1.0.

\begin{table}[t]
\setlength\tabcolsep{0.8pt} 
\caption{OOD detection performance for MNIST as the in-distribution dataset. AUC scores are shown. Half indicates HalfMNIST dataset, and Chimera indicates ChimeraMNIST dataset.}
\label{tb:MNIST-OOD}
\begin{center}
\begin{footnotesize}
\begin{sc}
\begin{tabular}{lccccc}
\toprule
 & ConstantGray & Half & Chimera & Omniglot  & Noise \\
\midrule
NAE    & \textbf{1.00} & \textbf{.999} & \textbf{.992} & \textbf{.995} & 1.00 \\
AE     & .934 & .523 & .761 & .885 & 1.00\\
PXCNN  & .982 & .154 & .334 & .665 & 1.00 \\
Glow   & \textbf{1.00} & .948 & .445 & .653 & 1.00  \\
IGEBM  & \textbf{1.00} & .986 & .948 & .946 & 1.00 \\
\bottomrule
\end{tabular}
\end{sc}
\end{footnotesize}
\end{center}
\vskip -0.1in
\end{table}

\subsection{Sample Generation}

\begin{figure}[h]
    \centering
    \includegraphics[width=0.48\textwidth]{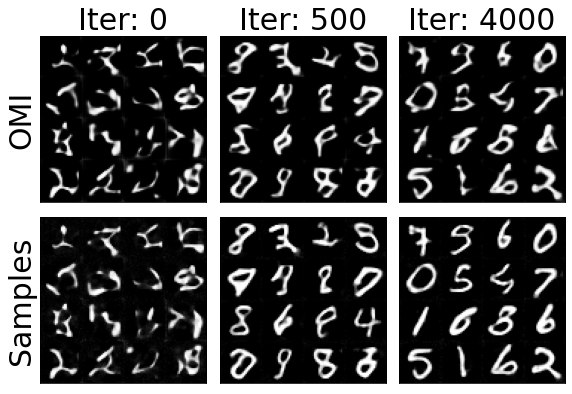}
    \caption{Samples collected from NAE during training. Initial states given by OMI and final states after MCMC on $\mathcal{X}$ are shown. Each column corresponds to a training iteration of NAE. Iter 0 is right after the autoencoder pre-training. NAE is trained on MNIST excluding the digit 9.}
    \label{fig:samples-along-training}
\end{figure}

\textbf{Visualization of Negative Samples}
We collect negative samples used in training NAE on MNIST and visualize them in Figure \ref{fig:samples-along-training}.
Additionally, OMI samples generated from the latent chain are also visualized.

Samples obtained right after pre-training (``Iter:0" in the figure) does not visually close to digits. Since sampling procedure tends to generates images with high $p_\theta(\bx)$, i.e., log $l_\theta(\bx)$, these samples may have lower reconstruction error than in-distribution MNIST digits.
As training proceeds, the samples become more visually similar to digits in MNIST.

\begin{figure}
    \centering
    \includegraphics[width=0.48\textwidth]{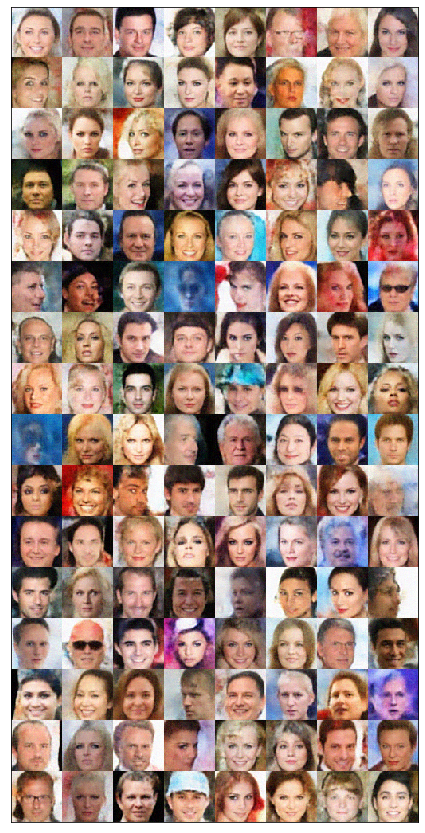}
    \caption{More samples from NAE trained on CelebA 64$\times$64. While most of the samples are visually sensible, a few generation failures can be spotted. Improving the sample generation process will be able to eliminate such non-realistic images.}
    \label{fig:more-celeba-samples}
\end{figure}

\textbf{Quantitative Result for Sample Quality}
We generate 50,000 samples from NAE trained on CelebA 64$\times$64 and compute FID score \cite{heusel2017gans}. We also visualize some of images generated by NAE in Figure \ref{fig:more-celeba-samples}. 
While FID score of NAE was not as low as ones from models specialized in generation, such as NCSN \cite{song2020improved}, FID score of NAE resides in a ballpark of what is achievable by autoencoder-based methods.
We believe that tuning network architecture and sampling procedure will result in enhanced samples.

\begin{table}[t]
\setlength\tabcolsep{1.5pt} 
\caption{FID score of 50,000 images generated from a model trained on CelebA 64$\times$64. A low FID score indicates that the generated images have similar statistics to the real images in Inception network's feature space.}
\label{tb:FID}
\begin{center}
\begin{footnotesize}
\begin{sc}
\begin{tabular}{lc}
\toprule
Model & FID \\
\midrule
NAE    & 94.00 \\
\midrule
From \citet{Ghosh2020From} \\
AE     &  127.85 \\
AE-L2     &  346.29 \\
VAE     &  48.12 \\
RAE-GP     &  116.30 \\
RAE-L2     &  51.13 \\
RAE-SN     &  44.74 \\
\midrule
From \citet{song2020improved} \\
NCSN (w/ denoising) & 26.89 \\
NCSNv2 (w/ denoising) & 10.23 \\
\bottomrule
\end{tabular}
\end{sc}
\end{footnotesize}
\end{center}
\vskip -0.1in
\end{table}

\section{Additional Discussion}

\subsection{Parameter Sensitivity Study}

To demonstrate the contribution of learning techniques and components used in NAE, we alter or ablate components in NAE and observe how it behaves differently.
As a model scenario, we consider a setting where digit 0-8 in MNIST are the inlier class and MNIST 9 is the outlier class.
To run multiple experiments, we shorten the number of NAE epochs to 20 for these experiments.

\textbf{The Effect of Pre-training}
We alter the number of training epochs in pre-training and observe how AUC of detection changes. 
The results are shown in Table \ref{tb:pre-training}.
Training as NAE is beneficial for the outlier detection, since AUC scores always increase after NAE training.
The gain from the pre-training starts to saturate after a certain number (50 in Table \ref{tb:pre-training}) of epochs, and the resulting performance remains within a certain range.
Therefore, significant performance boost can be obtained with the limited amount of pre-training.

\begin{table}[h]
\caption{The effect of pre-training as a conventional autoencoder. We vary the number of pre-training epochs and observe AUC scores before and after NAE training. Note that ``AUC Before NAE" indicates the outlier detection performance as a conventional autoencoder.}
\label{tb:pre-training}
\begin{center}
\begin{footnotesize}
\begin{sc}
\begin{tabular}{ccc}
\toprule
Pre-training & AUC & AUC\\
Epochs & Before NAE & After NAE  \\
\midrule
0 & .486  & .552  \\
10  & .581 & .754 \\
50  & .539 & .945 \\
100 & .532 & .937 \\
150 & .548 & .914 \\
200 & .537 & .919 \\
\bottomrule
\end{tabular}
\end{sc}
\end{footnotesize}
\end{center}
\vskip -0.1in
\end{table}

\textbf{The Effect of Sampling Parameters}
We mainly investigate the effect of three components: latent LMC chain, sample replay buffer, and spherical latent space.
Table \ref{tb:ablation} shows different configurations experimented and their anomaly detection performance in AUC. Experiment No.1 represents the original NAE configuration used in the experiments in the main manuscript. 

\begin{table}[h]
\small
\setlength\tabcolsep{3pt} 
  \caption{Experiments with sampling parameters.}
  \label{tb:ablation}
  \centering
  \begin{tabular}{ccccc}
  \toprule
\multirow{ 2}{*}{No.} &Latent chain &Using sample & Using spherical & \multirow{ 2}{*}{AUC}  \\
&length&  replay buffer&	 latent space 	& \\
\midrule
1&	10&Yes&	Yes& .957\\
2&	\textbf{\underline{0}}&	\textbf{\underline{No}}&	Yes&.839\\
3&	\textbf{\underline{50}}&Yes&	Yes&.944\\
4&	10&\textbf{\underline{No}}&	Yes&.936\\
5&	\textbf{\underline{50}}&	\textbf{\underline{No}}&	Yes&.952\\
6&	10&	Yes&	\textbf{\underline{No}}&.662\\
    \bottomrule
  \end{tabular}
\end{table}

\begin{itemize}[leftmargin=1em,topsep=0pt,itemsep=0pt]
\item The use of latent chain is substantially beneficial, as shown from the comparison between No.1 and No.2.
\item A longer latent chain is needed for a better result. Comparing No.2, No.4, and No.5, the performance increases monotonically with the chain length. Note that longer chain length requires proportionally longer time in training. 
\item The sample replay buffer helps to reduce the length of the latent chain required in training. We see that with the sample replay buffer, the latent chain with 10 steps (No.1) can achieve performance comparable to that of 50 latent chain steps (No.5). In other words, using sample replay buffer may reduce the training time up to five times without the loss of performance.
\item The gain from a longer latent chain saturates. Especially when a sample replay buffer is used, a chain longer than a certain threshold does not improve the performance (No.1 \& No.3).
\item The use of spherical latent space is essential in achieving a good performance. 
\end{itemize}

\textbf{The Effect of $D_{\bz}$} 
One of the most critical hyperparameter in autoencoders is the size of the latent space. We vary $D_\bz$ from 2 to 256 and observe how the performance of detecting digit 9 varies accordingly. The result is shown in Table \ref{tb:zdim}

The performance of an autoencoder is very sensitive to the choice of $D_\bz$, achieving good performance only in the vicinity of $D_\bz=8$. Meanwhile, NAE achieves better performance than NAE in a wider range of $D_\bz$, from 16 to 128. Therefore, NAE is more robust to the choice of $D_\bz$ than an autoencoder.

\begin{table}[ht]
\caption{The effect of the latent dimensionality $D_\bz$. The best performance from an autoencoder is underlined, and the performance of NAE better than the best performance of an autoencoder is marked as bold.}
\label{tb:zdim}
\begin{center}
\begin{footnotesize}
\begin{sc}
\begin{tabular}{ccc}
\toprule
$D_\bz$ & AE AUC & NAE AUC\\
\midrule
2 &  .626&   .591  \\
4  &  .748& .550 \\
8  &  \underline{.825}&  .723\\
16 &  .614&  \textbf{.911}\\
32 &  .525&  \textbf{.926}\\
64 &  .506&  \textbf{.884}\\
128 &  .546& \textbf{.830}\\
256&   .556& .732\\
\bottomrule
\end{tabular}
\end{sc}
\end{footnotesize}
\end{center}
\vskip -0.1in
\end{table}

\subsection{Computational Properties of NAE}

The inference step of NAE is as fast as that of a conventional autoencoder, requiring no more than a single forward pass. The Langevin Monte Carlo (LMC) is the most time consuming procedure in NAE training and sampling
A single draw of sample involves dozens of LMC transition steps, and each transition step requires multiple backward passes one of which is comparable to a single training step for an autoencoder.
However, the number of transition step required is less than 100 per step, and it is feasible to run it on a modern hardware.
Thanks to availability of the pre-training, NAE becomes a decent sampler as well as an outlier detector even after a few epochs, while the performance tends to gradually increase for a few dozens of epochs.
A single epoch of NAE training on MNIST with Conv28 network takes 10 minutes using a single Tesla V100 GPU.
A single epoch of NAE training on CIFAR-10 with Conv32 network takes 500 seconds.

\subsection{Outlier Reconstruction}
\label{sec:outlier-recon}

\begin{figure}[ht]
  \begin{center}
    \includegraphics[width=0.4\textwidth]{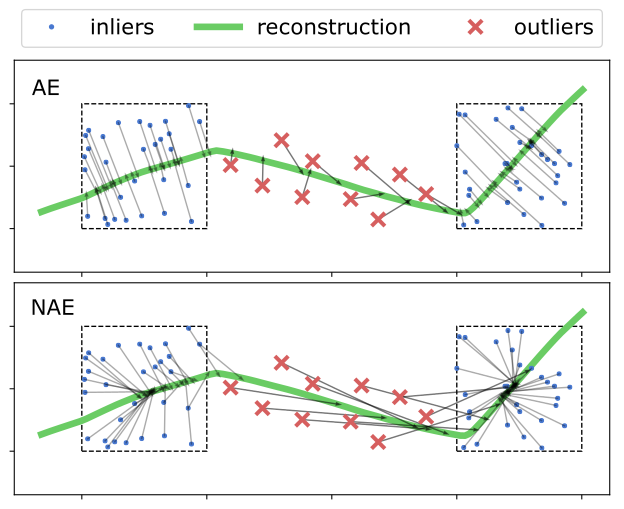}
  \end{center}
  \caption{AE and NAE trained on a bi-modal distribution. Here, NAE is trained with its decoder fixed. The green lines denotes the decoder manifolds. The dotted lines link inputs and their reconstructions.}
  \label{fig:2d-cluster}
\end{figure}

The outlier reconstruction is a phenomenon that an autoencoder unexpectedly succeeds in reconstructing an input even though it is located outside of the training distribution.
In this section, we provide illustrative examples that show that outlier reconstruction is a consequence from the inductive biases of an autoencoder.

\textbf{Multi-modal data} When the training data distribution consists of multiple clusters, the outliers from the region between the clusters are likely to be reconstructed.
Figure \ref{fig:2d-cluster} depicts 2D synthetic data generated from a mixture of two disconnected uniform distributions and their reconstruction from autoencoders with one-dimensional latent space.
The outliers (red crosses) from the middle of two clusters show reconstruction errors (the length of thin black lines) smaller than some inliers (blue dots).
\cite{tong2019fixing} noted this type of outlier reconstruction and mentioned that outliers ``close to the mean" of data or ``in the convex hull" of data are likely to be reconstructed.

This phenomenon arises from the inductive bias of an autoencoder that its encoder and decoder are smooth mappings.
The extreme case of this inductive bias can be found in linear principal component analysis (PCA). PCA, a linear counterpart of an autoencoder \citep{bourlard1988auto}, would reconstruct any outliers which reside on the principal axis.
Note that this phenomenon is consistent with the objective function of an autoencoder and PCA, as the objective does not penalize the reconstruction of outliers.




\textbf{Compositionality} When there is a compositional structure in data, we can still observe a reconstructed outlier even if it lies outside of the convex hull of training data.
The data are compositional if each datum can be broken down into smaller reusable components; For example, MNIST can be considered highly compositional, since a digit image can be decomposed into smaller sub-patterns, such as straight lines and curves.
An outlier can be successfully reconstructed when composed of a subset of components existing in the training data.

HalfMNIST and ChimeraMNIST datasets (See Section \ref{sec:datasets}) are constructed to demonstrate the effect of compositionality in outlier reconstruction.
Although these images are not in the convex hull of MNIST digits, they share components found in MNIST.
As shown in Figure \ref{fig:mnist_recon}, an autoencoder trained on MNIST have no problem reconstructing them and achieves poor AUC scores in classifying HalfMNIST and ChimeraMNIST from MNIST (See Table \ref{tb:MNIST-OOD}).

It seems that an autoencoder learns to reconstruct each part of an image separately but is not able to judge whether the combination of the parts is valid as a whole.
This compositional way of processing facilitates generalization of a model \citep{keysers2019measuring}, but the generalization of reconstruction in OOD inputs is not desirable for an autoencoder-based outlier detector.

\begin{figure}[h]
  \begin{center}
    \includegraphics[width=0.5\textwidth]{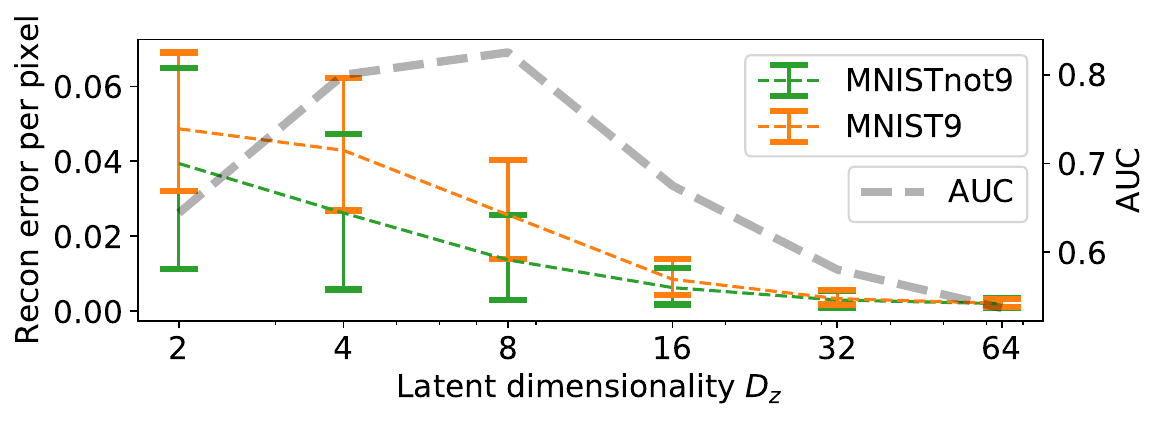}
  \end{center}
  \caption{Detecting hold-out digit 9 from the rest of MNIST. Reconstruction errors and AUC scores are shown across multiple values of $D_\bz$. The error bars denote 80-percentile around the means.}
  \label{fig:recon-vs-dim}
\end{figure}

\textbf{Distributed representation} We suspect the outlier reconstruction due to compositional processing may be attributed to the distributed representation \citep{mikolov2013distributed} used in an autoencoder.
To show the effect of the distributed representation, we train autoencoders on MNIST with the digit 9 excluded (MNISTnot9) and measure the reconstruction error of the digit 9 (MNIST9) under multiple values of latent dimensionality $D_\bz$.
Figure \ref{fig:recon-vs-dim} shows the result.
We observe the outlier reconstruction of 9 possibly due to the compositional processing mentioned above. However, the outlier reconstruction occurs only when $D_\bz$ is large.
The latent representation is more distribution for large $D_\bz$, as a larger number of hidden neurons are used to represent an input.
This observation suggests that the distributed representation used in an autoencoder enables the compositional processing and thus facilitates outlier reconstruction.

\section{Experimental Detail}
Here, we present details on our experiments, including datasets, network architectures, and details regarding implementing NAE and baselines.
A sample code is provided in the supplementary material.
We will make the whole experiment codebase public.

\subsection{Datasets} \label{sec:datasets}

Here, we present the details on the datasets used in the experiments.

\textbf{2D Synthetic Dataset} We use a two-dimensional mixture of 8 Gaussians initially used in \cite{grathwohl2018ffjord}.
The means of 8 Gaussians are $[2\sqrt{2},0]^\top$, $[0,2\sqrt{2}]^\top$, $[-2\sqrt{2},0]^\top$, $[0,-2\sqrt{2}]^\top$, $[2,2]^\top$, $[-2,2]^\top$, $[2,-2]^\top$, $[-2,-2]^\top$. All Gaussians are isotropic and have a covariance matrix of $\frac{\sqrt{2}}{4}\mathbf{I}$.

\begin{table*}[h]
\small
  \caption{Summary of dataset statistics.}
  \label{tb:dataset}
  \centering
  \begin{tabular}{ccccc}
    \toprule
    Dataset & Original Shape & Training  & Validation & Test  \\
    \midrule
    Constant (Synthetic) & - & -  & 4,000  & 4,000 \\
    MNIST & 1$\times$28$\times$28  & 54,000 & 6,000  & 10,000  \\
    FashionMNIST & 1$\times$28$\times$28  & 54,000   & 6,000  & 10,000  \\
    Omniglot & 1$\times$28$\times$28 & - & - & 13,180 \\
    SVHN & 3$\times$32$\times$32  & 65,930  & 7,327  & 20,632  \\
    CIFAR-10 & 3$\times$32$\times$32  & 45,000   & 5,000  & 10,000 \\
    CelebA & 3$\times$178$\times$218  & 162,079  & 20,260  & 20,260  \\
    ImageNet32 & 3$\times$32$\times$32 &  1,024,919  &  256229    &   49999      \\
    Noise (Synthetic) & 3$\times$32$\times$32  & -  & 4,000  & 4,000 \\
    HalfMNIST (Synthetic) & 1$\times$28$\times$28  & -  & -  & 10,000 \\
    ChimeraMNIST (Synthetic) & 1$\times$28$\times$28  & -  & -  & 10,000 \\
    \bottomrule
  \end{tabular}
\end{table*}

\textbf{Image Dataset}
Table \ref{tb:dataset} summarizes image datasets used in our experiment.
Pixel values are de-quantized by adding uniform noise $u\sim U(0,1)$ and then normalized to interval $[0,1]$.

For MNIST, FashionMNIST, SVHN, and CIFAR-10, we use the predefined test splits as the test sets and randomly select 10\% from the train splits as validation set. 

We use the non-background Omniglot dataset only during the test phase. Omniglot images are inverted in order to produce black background.

For CelebA, we used the official train-validation-test split. Each image in CelebA is center-cropped in 140$\times$140 and then resized into 32$\times$32 or 64$\times$64 depending on the experiment. When MNIST and Fashion MNIST are needed to be fed to a network trained on 32$\times$32 images, an input is resized into 32$\times$32.
The training set of ImageNet32 \citep{oord2016pixel} is the randomly selected 80\% from the original train split, and the rest 20\% are used as the validation set. We use the original validation split as the test set in our experiment.

ConstantGray dataset is a synthetic dataset with images all pixels of which have the same gray-scale value. 
An image in ConstantGray dataset is a 1$\times$28$\times$28 or 3$\times$32$\times$32 array filled with a single number drawn from the interval $[0, 1]$.
As shown in our experiment, most of existing autoencoders tend to reconstruct an image from ConstantGray with significantly low reconstruction error. 
Similarly, generative models tend to assign a very high likelihood to an image from ConstantGray. A similar observation is provided in \citet{Serra2020Input}.

Noise dataset is created by randomly generating each pixel value from interval $[0,1]$.
Noise dataset is very difficult to reconstruct exactly.
Classifying Noise dataset can be a sanity check for an outlier detection that it can detect very obvious outliers.

Two additional datasets synthesized from MNIST are used. HalfMNIST dataset is created by randomly replacing the upper or the lower half of an MNIST test image with zero pixels. 
An image in ChimeraMNIST dataset is constructed by randomly joining the upper and the lower part of two MNIST test images from different classes. We add six zero pixels in the middle as a padding.
HalfMNIST and ChimeraMNIST images are clearly OOD from MNIST but partially shares local visual features.

\subsection{Network Architecture}

\begin{table*}[t]
\small
  \caption{Network architectures for autoencoders.}
  \label{tb:arch}
  \centering
  \begin{tabular}{cccc}
  \toprule
& FCRes2 ($D_\bx$=2) & Conv28 ($D_\bx=$1$\times$28$\times$28) & Conv32 ($D_\bx=$3$\times$32$\times$32) \\
\midrule
Encoder &
  \begin{tabular}{@{}l@{}}
  FC(2, 256)-\\
  FCRes(256, 1024)-\\
  FCRes(256, 1024)-\\
  FCRes(256, 1024)-\\
  FCRes(256, 1024)-\\
  FCRes(256, 1024)-ReLU-\\
  FC(256, $D_\bz$)
  \end{tabular}
&
 \begin{tabular}{@{}l@{}}
 Conv2d(1, 32, 3, 1, 1)-ReLU- \\ 
 Conv2d(32, 64, 3, 1, 1)-ReLU-MaxPool(2)- \\
 Conv2d(64, 64, 3, 1, 1)-ReLU- \\
 Conv2d(64, 128, 3, 1, 1)-ReLU-MaxPool(2)- \\
 Conv2d(128, 1024, 4, 1, 0)-ReLU- \\
 FC(1024, $D_{\bz}$)
 \end{tabular}
& 
\begin{tabular}{@{}l@{}}
 Conv2d(3, 32, 4, 2, 0)-ReLU- \\ 
 Conv2d(32, 64, 4, 2, 0)-ReLU- \\
 Conv2d(64, 128, 4, 2, 0)-ReLU- \\
 Conv2d(128, 256, 2, 2, 0)-ReLU- \\
 Conv2d(256, $D_{\bz}$, 1, 1, 0)-ReLU- \\
 ResAtt($D_{\bz}$, 1024, $D_{\bz}$)
 \end{tabular}
\\
\midrule
Decoder & 
  \begin{tabular}{@{}l@{}}
  FC($D_\bz$, 256)-\\
  FCRes(256, 1024)-\\
  FCRes(256, 1024)-\\
  FCRes(256, 1024)-\\
  FCRes(256, 1024)-\\
  FCRes(256, 1024)-ReLU-\\
  FC(256, 2)
  \end{tabular}
&
\begin{tabular}{@{}l@{}}
ConvT2d($D_{\bz}$, 128, 4, 1, 0)-ReLU-Bilinear(2)- \\ 
ConvT2d(128, 64, 3, 1)-ReLU- \\
ConvT2d(64, 64, 3, 1)-ReLU-Bilinear(2)- \\
ConvT2d(64, 32, 3, 1)-ReLU- \\
ConvT2d(32, 1, 4, 1)-Sigmoid
\end{tabular}
& 
\begin{tabular}{@{}l@{}}
ConvT2d($D_{\bz}$, 256, 6, 1)-ReLU- \\ 
ConvT2d(256, 128, 4, 2)-ReLU- \\
ConvT2d(128, 64, 4, 2)-ReLU- \\
ConvT2d(64, 3, 3, 1)-Sigmoid
\end{tabular}
\\
    \bottomrule
  \end{tabular}

    \begin{tabular}{ccc}
  \toprule
& Conv32Big ($D_\bx=$3$\times$32$\times$32) & Conv64 ($D_\bx=$3$\times$64$\times$64) \\
\midrule
Encoder 
&
 \begin{tabular}{@{}l@{}}
 Conv2d(3, 128, 3, 1, 1)- \\
 Res(128, 128, 128)-AvgPool(2)- \\ 
 Res(128, 128, 128)- \\ 
 Res(128, 256, 256)-AvgPool(2)- \\ 
 Res(256, 256, 256)- \\ 
 Res(256, 256, 256)-AvgPool(2)- \\ 
 Res(256, 256, 256)-LReLU- \\ 
 AvgPool(4)-FC(256, $D_{\bz}$)
 \end{tabular}
& 
\begin{tabular}{@{}l@{}}
 Conv2d(3, 256, 5, 2, 0)-ReLU- \\ 
 Conv2d(245, 512, 5, 2, 0)-ReLU- \\
 Conv2d(512, 1024, 5, 2, 0)-ReLU- \\
 Conv2d(1024, 2048, 5, 2, 0)-ReLU- \\
 FC(2048, $D_{\bz}$) \\
 \end{tabular}
\\
\midrule
Decoder
&
\begin{tabular}{@{}l@{}}
ConvT2d($D_\bz$, 128, 4, 1)-\\
ResUp(128, 128, 128, 2)-\\
ResUp(128, 128, 128, 2)-\\
ResUp(128, 128, 128, 2)-ReLU-\\
Conv2d(128, 3, 3, 1, 1)-Sigmoid\\
\end{tabular}
& 
\begin{tabular}{@{}l@{}}
ConvT2d($D_{\bz}$, 1024, 8, 1)-ReLU- \\ 
ConvT2d(1024, 512, 4, 2, 1)-ReLU- \\
ConvT2d(512, 256, 4, 2, 1)-ReLU- \\
ConvT2d(256, 128, 4, 2, 1)-ReLU- \\
ConvT2d(128, 3, 1, 1, 0)-Sigmoid
\end{tabular}
\\
    \bottomrule
  \end{tabular}
\end{table*}

Several network architectures are used throughout the experiments, and the architectures are summarized in Table \ref{tb:arch}.
We name the architectures as \textbf{FCRes2}, \textbf{Conv28}, \textbf{Conv32}, \textbf{Conv32Big}, and \textbf{Conv64}. The numbers indicate the dimensionality of an input or a spatial dimension of an input image.
All networks do not have batch normalization, as batch normalization violates the assumption in EBM that the energy is a deterministic function of an input.
Conv32 and Conv64 architectures are similar to ones used in \cite{Ghosh2020From}.
The encoder of Conv32Big is the energy function used in \citet{du2019}, and the decoder of Conv32big is the generator used in \citet{miyato2018spectral}.

FCRes2 is used in 2D density estimation.
Conv28 is used in the experiments with MNIST dataset.
Conv32 is used in the experiments with CIFAR-10 dataset, and Conv32Big is used in the experiments with ImageNet 32$\times$32 dataset.
Conv64 is used in the experiments with CelebA 64$\times$64.

We use the following notations to denote operations in networks.

\begin{itemize}[leftmargin=1em,topsep=0pt,itemsep=-1pt]
\item Conv2d(\texttt{in}, \texttt{out}, \texttt{kernel}, \texttt{stride}, \texttt{padding}): A 2D convolutional operation with bias.
\item ConvT2d(\texttt{in}, \texttt{out}, \texttt{kernel}, \texttt{stride}, \texttt{padding}): A 2D transposed convolutional operation with bias.
\item ReLU: A rectified linear unit activation function. $y=\max\{0, x\}$.
\item LReLU: A leaky ReLU function with the negative slope of 0.2.
\item Sigmoid: A sigmoid activation function. $y=1/(1+\exp(-x))$.
\item MaxPool(\texttt{kernel\_size}): A max-pooling operator with the window of \texttt{kernel\_size}$\times$\texttt{kernel\_size} and the stride of \texttt{kernel\_size}.
\item Bilinear(\texttt{scale\_factor}): A bilinear upscaling operation that upsamples an input array into an \texttt{scale\_factor} times larger array.
\item Nearest(\texttt{scale\_factor}: A nearest upscaling opteration which upsamples an input array into an \texttt{scale\_factor} times larger array.
\item FC(\texttt{in}, \texttt{out}): A fully-connected layer performing a linear mapping with a bias.
\item FCRes(\texttt{in}, \texttt{hidden}, \texttt{out}): A residual block composed of two fully-connected layers.
\begin{align*}
F_1=&\text{FC(\texttt{in}, \texttt{hidden})} \\
F_2=&\text{FC(\texttt{hidden}, \texttt{out})} \\
y=&x + F_2(\text{ReLU}(F_1(\text{ReLU}(x))))
\end{align*}
for an input $x$ and an output $y$.
\item Res(\texttt{in}, \texttt{hidden}, \texttt{out}): A convolutional residual block composed of two \texttt{Conv2d} layers. For an input $x$ and an output $y$,
\begin{align*}
F_1&=\text{Conv2d(\texttt{in}, \texttt{hidden}, 3, 1, 1)}\\ F_2&=\text{Conv2d(\texttt{hidden}, \texttt{out}, 3, 1, 1)}\\
y&=x + F_2(\text{LReLU}(F_1(\text{LReLU}(x))))
\end{align*}
\item ResUp(\texttt{in}, \texttt{hidden}, \texttt{out}, \texttt{scale}): A residual block with upsampling. 
\begin{align*}
F_1&=\text{Conv2d(\texttt{in}, \texttt{hidden}, 3, 1, 1)}\\ F_2&=\text{Conv2d(\texttt{hidden}, \texttt{out}, 3, 1, 1)}\\
U&=\text{Nearest}(\texttt{scale})\\
y&=U(x) + F_2(\text{ReLU}(F_1(U(\text{ReLU}(x)))))
\end{align*}
\item ResAtt(\texttt{in}, \texttt{hidden}, \texttt{out}): A residual attention block which has two residual blocks $R_1=\text{Res(\texttt{in}, \texttt{hidden}, \texttt{out})}$ and $R_2=\text{Res(\texttt{in}, \texttt{hidden}, \texttt{out})}$ and operates as $y = x + R_1(x) \odot \text{Sigmoid}(R_2(x)) $, where $\odot$ denotes an element-wise product.
\end{itemize}

\subsection{NAE Implementation} \label{sec:nae-implementation}

\begin{table*}[t]
\caption{Sampling parameters for NAE. See Section \ref{sec:nae-implementation} for description.}
\label{tb:sampling-hyperparameters}
\begin{center}
\begin{small}
\begin{sc}
\begin{tabular}{lcccc}
\toprule
 & 2D & MNIST & CIFAR-10 & CelebA  \\
\midrule
Latent space & $\RE^{D_\bz}$ & $\mathbb{S}^{D_\bz-1}$ & $\mathbb{S}^{D_\bz-1}$ & $\mathbb{S}^{D_\bz-1}$ \\
Step size $\lambda_\bz$ & 0.005 & 0.2 & 1 & 1 \\
Noise $\sigma_\bz$ & 0.1 & 0.05 & 0.02 & 0.02 \\
Latent chain length $\tau_\bz$    & 10 & 10 & 20 & 20 \\
Replay buffer for latent chain    & O & O & O & O\\
Step size $\lambda_\bx$     & 0.005 & 10 & 10 & 10\\
Noise $\sigma_\bx$      & 0.1 & 0.05 & 0.02 & 0.02\\
Main chain length $\tau_\bx$      & 30 & 50 & 40 & 40\\
$\sigma_\bx$ annealing   & X & O & O & O \\
Gradient clipping in $\mathcal{X}$ & X & O & O & O \\
Temperature $T=T_\bz$ & 0.5, trainable & 1, fixed  & 1, fixed  & 1, fixed  \\
MH rejection & O & X & X & X \\
\bottomrule
\end{tabular}
\end{sc}
\end{small}
\end{center}
\vskip -0.1in
\end{table*}

The procedure for generating a negative sample in NAE is described in Algorithm \ref{alg:omi}.

\textbf{Sampling Parameters and Techniques} The most important hyperparameter in NAE is the parameters related to MCMC sampling. 
Detailed MCMC parameters can be found in Table \ref{tb:sampling-hyperparameters}.

Note that the step size parameters and the noise parameters in 2D setting follows the theoretically motivated relationship $2\lambda = \sigma^2$.
For the rest of the cases, $2\lambda \neq \sigma^2$, meaning that the \emph{effective} temperatures are implicitly fixed to values different from 1.
In all experiments, we set $T=T_\bz$, i.e., the temperature of the latent chain is tied to the temperature of NAE.
In 2D dataset experiments, the temperature is tuned with gradient descent. For faster convergence, in updating $T$, we use a learning rate 100 times larger than that used to update the model parameters.

Following \citet{du2019}, we clip the gradient $\nabla_\bx \log E_\theta (\bx)$ is clipped at 0.01.

We use the sample replay buffer with the buffer size of 10,000 and the replay ratio of 95\% as described in \citet{du2019}. A starting point of a latent chain is drawn from the noise distribution $q_0(\bz)$ with the probability of 5\% or randomly drawn from the replay buffer with the probability of 95\%.

The choice of a noise distribution $q_0(\bz)$ is dependent on the configuration of the latent space. For $\RE^{D_\bz}$, $q_0(\bz)$ is a standard normal distribution. For $\mathbb{S}^{D_\bz -1}$, a uniform distribution is used. Samples from a uniform distribution on $\mathbb{S}^{D_\bz -1}$ is generated by first draw samples from a standard normal distribution in $\RE^{D_\bz}$, then project them onto $\mathbb{S}^{D_\bz -1}$ by dividing with its norm.

$\sigma_\bx$ annealing in Table \ref{tb:sampling-hyperparameters} indicates that during MCMC in $\mathcal{X}$, we let the standard deviation of $\epsilon_\bx$ decay as the visible chain proceeds:
$\text{std}(\epsilon_\bx)=0.05 / (1 + \text{step})$ for $\text{step}=0,\cdots,s$ with $s$ as the number of steps in $\mathcal{X}$.
This contributes to obtaining finer samples.

For generating new samples, a longer latent chain is used instead of a sample replay buffer. In our experiments, we let a latent chain be 8 times longer than the chain used in training.

\textbf{Learning of NAE} 
NAE is trained using Adam optimizer \cite{kingma2014adam} with the learning rate of $1\times10^{-5}$ for 50 epochs.
When using the reconstruction error as the energy function, we divide the reconstruction error by the dimensionality of an input data.
This keeps the energy in a reasonable scale and facilitates hyperparameter setting.

Autoencoder pre-training is not used in 2D experiments but is conducted for 100 epochs for MNIST and 120 epochs for CIFAR-10.

The encoders and the decoders in NAE are regularized as follows:
\begin{itemize}[leftmargin=1em,topsep=0pt,itemsep=-1pt]
    \item FCRes2: No regularization;
    \item Conv28: L2 norm of the encoder's weights with the coefficient 0.0001;
    \item Conv32: L2 norm of the encoder's weights with the coefficient 0.0001;
    \item Conv32Big: Spectral normalization \cite{miyato2018spectral} (encoder) and group normalization \cite{wu2018group} with 8 groups (decoder);
    \item Conv64: group normalization with 8 groups (encoder \& decoder).
\end{itemize}

\begin{algorithm}[tb]
   \caption{Negative sample generation using OMI}
   \label{alg:omi}
\begin{algorithmic}
   \STATE {\bfseries Input:} Sample replay buffer $\mathcal{B}$, noise distribution $q_0(\bz)$, latent energy $H_\theta(\bz)$, NAE energy $E_\theta(\bz)$, decoder $f_d(\bz)$
   \STATE \;\;\;\;Latent chain parameters $\sigma_\bz$, $\lambda_\bz$, $\tau_\bz$
   \STATE \;\;\;\;Main chain parameters $\sigma_\bx$, $\lambda_\bx$, $\tau_\bx$
   \STATE // Initialization
   \STATE Draw $u$ from $Uniform(0,1)$. 
   \IF{$u < 0.95$}
   \STATE Draw $\bz_0$ from $\mathcal{B}$
   \ELSE
   \STATE Draw $\bz_0$ from $q_0(\bz)$
   \ENDIF
   \STATE // Latent chain
   \FOR{$t=0$ {\bfseries to} $\tau_\bz$}
   \STATE $\bz_t = -\lambda_\bz\nabla H_\theta(\bz_{t-1}) +\sigma_\bz \epsilon$, $\epsilon\sim\mathcal{N}(\mathbf{0}_\bz,\mathbf{I}_\bz)$
   \STATE // Project $\bz_t$ to $\mathbb{S}^{D_\bz-1}$ if needed
   \ENDFOR
   \STATE Append $\bz_{\tau_\bz}$ to $\mathcal{B}$
   \STATE $\bx_0=f_d(\bz_{\tau_\bz})$ // On-manifold initialization
   \STATE // Main chain
   \FOR{$t=0$ {\bfseries to} $\tau_\bx$}
   \STATE $\bx_t = -\lambda_\bx\nabla E_\theta(\bx_{t-1}) +\sigma_\bx \epsilon$, $\epsilon\sim\mathcal{N}(\mathbf{0}_\bx,\mathbf{I}_\bx)$
   \STATE // Metropolis-Hastings rejection step if needed
   \STATE // Anneal $\sigma_\bx$ if needed
   \ENDFOR
   \STATE {\bfseries return} $\bx_{\tau_\bx}$ // Negative sample
   
\end{algorithmic}
\end{algorithm}

\subsection{Baseline Implementation}

In all autoencoder-based methods, i.e., NAE, AE, DAE, WAE, VAE, DAGMM, VQVAE and Ganomaly, we ensure that they all use the same network architecture.

The optimal $D_\bz$ is searched among $\{2, 4, 8, 16, 32, 64, 128, 256\}$. The best parameter is selected according to AUC of classifying a separate OOD dataset, as described in the main manuscript.

PixelCNN++ \cite{salimans2017pixelcnn++} is implemented based on an open-source code base\footnote{\url{https://github.com/pclucas14/pixel-cnn-pp}}. We use set parameters as \texttt{nr\_resnet}=5 and \texttt{nr\_filters}=80.
Input values are scales to $[-1, 1]$ only when running PixelCNN++.

Glow \cite{kingma2018glow} is also implemented based on the open-source  repository\footnote{\url{https://github.com/chaiyujin/glow-pytorch}}.
We set $K=12$, $L=1$, $\texttt{hidden\_channels}=64$. 

We failed to obtain sensible result using GPND \cite{pidhorskyi2018generative}\footnote{\url{https://github.com/podgorskiy/GPND}}.

DAGMM \cite{zong2018deep} is implemented based on the public repository\footnote{\url{https://github.com/danieltan07/dagmm}}. We failed to train it on CIFAR-10 such that it produces AUC of 1.0 for Noise dataset, and therefore we exclude the result from the Table 2 of the main manuscript.

WAE-MMD is implemented. We use median heuristic to determine the kernel parameter of RBF kernel. Regularization coefficient is searched between 0.001 and 0.1

For DAE, we use Gaussian noise with standard deviation of 0.3.

We implement VQVAE based on the PyTorch version\footnote{\url{https://github.com/ritheshkumar95/pytorch-vqvae}}.
In the reproduction of Ganomaly, we use the official implementation from the authors\footnote{\url{https://github.com/samet-akcay/ganomaly}}.

\end{document}